\documentclass[11pt]{article}

\usepackage{agenticlearning-xelatex}

\usepackage{algorithm}
\usepackage{algpseudocode}

\shorttitle{Conceptual Creativity as Meta-Learning}
\shortauthor{Ren}

\title{Seeking the Unfamiliar but Memorable:\\
Conceptual Creativity as Meta-Learning}
\author{
  Mengye Ren \\
  Agentic Learning AI Lab, New York University \\
  \texttt{mengye@nyu.edu} \\
  \url{https://agenticlearning.ai/conceptual-creativity-meta-learning/}
}
\date{}

\begin{document}

\maketitle
{\makeatletter\renewcommand{\@makefntext}[1]{\noindent\raggedright #1}\makeatother\footnotetext{Published at the 40th Conference on Neural Information Processing Systems (NeurIPS 2026).}}

\begin{abstract}
What does it mean to create a new concept, rather than retrieve a familiar one? Repeatedly sampling a generative model at the same prompt produces variations with similar styles and typical content. We propose that creativity is the production of stimuli that are unfamiliar to an adaptive observer at first sight, but quickly learnable from a few exposures. We formalize this as a \textit{Creator-Appraiser} pair: a Creator generates a candidate, an Appraiser adapts to it for a few inner-loop learning steps, and the Appraiser's improvement becomes the reward the Creator optimizes through. We instantiate the framework with diffusion as the Creator, an autoencoder Appraiser on MNIST, and a CLIP Appraiser with a low-rank adapter for natural images. With a frozen diffusion model and no additional language conditioning, the meta-learning gradient is enough to produce both stylistic variations and concept compositions beyond what the base model can generate on its own.
\end{abstract}

\section{Introduction}

Humans recognize new concepts from just one or a few examples. Walking into a museum and encountering a Monet painting is enough to begin grasping the ``concept'' of Impressionism and to recognize its style in other works. This capacity for few-shot concept learning is a cornerstone of intelligent behavior~\citep{fei2006one,lake2011one}, and its algorithmic counterparts have advanced rapidly through meta-learning~\citep{finn2017model}, prototypical representations~\citep{snell2017prototypical}, and in-context learning~\citep{brown2020language}. But this raises a more fundamental question: \emph{what does it mean to create a new concept in the first place?} While humans and machines can readily learn concepts like Impressionism, the creative act itself remains elusive.

Before defining creation, it helps to see that not all novelty is meaningful. An infinite stream of white-noise images is distinct sample by sample, yet each is perceptually indistinguishable from the next and offers no structure to grasp. Conversely, a line, a square, or the sequence $1, 2, 3, \ldots$ is entirely comprehensible, yet lacks the novelty of a creative act. A concept worth creating sits between these extremes. This matches the standard definition in the creativity literature, which asks that a creative product be both novel and valuable~\citep{boden2004creative,runco2012standard,ritchie2007some}. For a new concept to be valuable, at least from a perceptual perspective, it must be grounded in our existing representations of the world; it must, on some level, be understandable from past experience.

One path to building creative artifacts is through novel composition of existing parts. A mermaid is imagined as a human upper body fused with a fish tail; the verbal recipe makes the creation easily described and learned. Composition, however, is not the only path. Some concepts are perceived as monolithic: the color fields of Mark Rothko, the iconic blue of Yves Klein, or the splash patterns of Jackson Pollock are not naturally read as combinations of simpler parts. A theory of creativity must accommodate both, and the difference may lie in which layers of an information-processing system are engaged.

\begin{figure}[t]
\centering
\begin{subfigure}[b]{0.3\textwidth}
  \centering
  \includegraphics[width=0.95\textwidth]{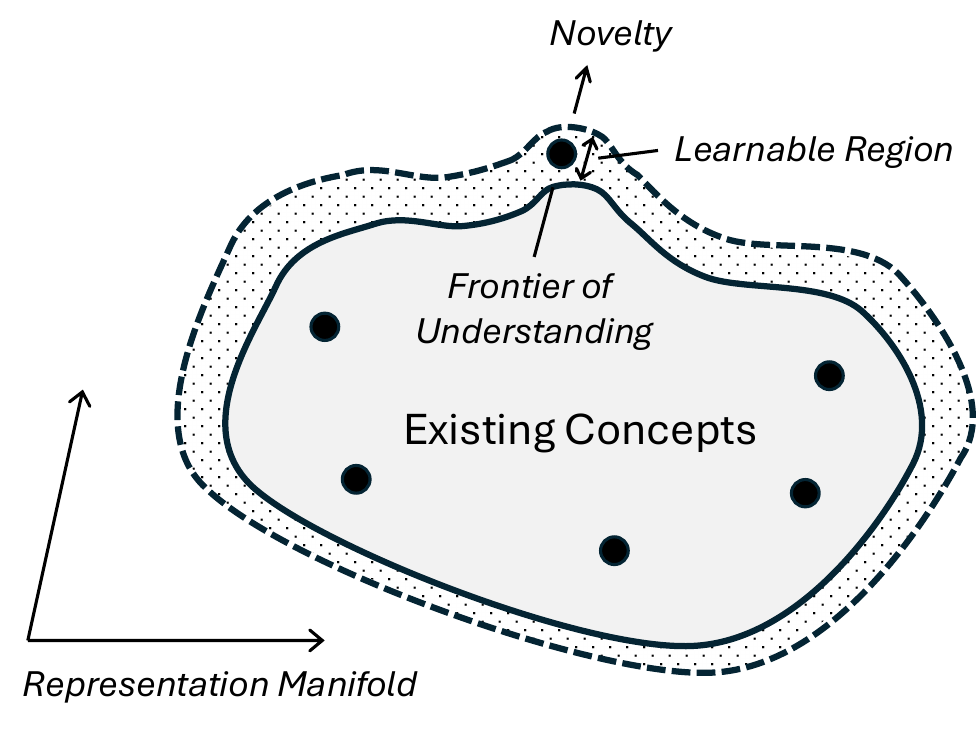}
  \caption{Frontier of understanding}
  \label{fig:intro}
\end{subfigure}
\hfill
\begin{subfigure}[b]{0.58\textwidth}
  \centering
  \includegraphics[width=0.95\textwidth]{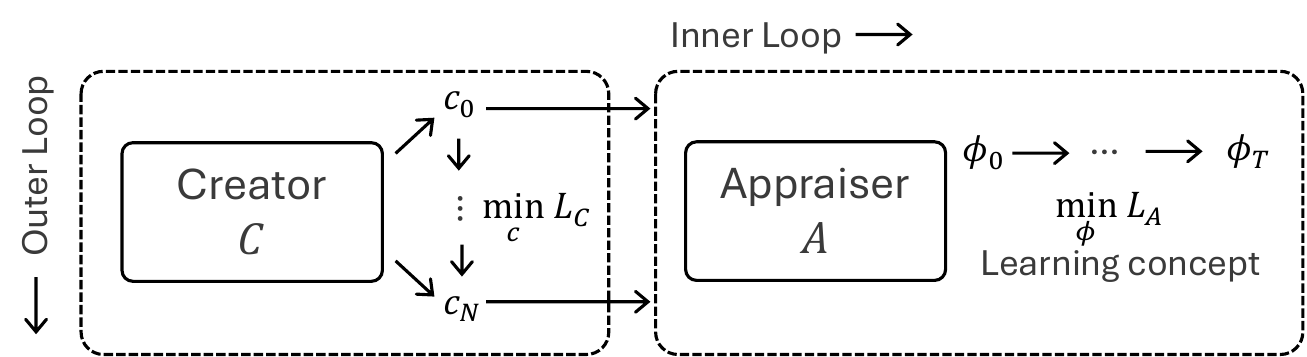}
  \caption{Creator-Appraiser Meta-Learning.}
  \label{fig:creator_appraiser}
\end{subfigure}
\caption{\textbf{(a)} Creativity happens at the frontier of understanding formed by existing concepts on the representation manifold learned from past experience. \textbf{(b)} The Creator-Appraiser framework, with a Creator $C$ and an Appraiser $A$. $A$ learns the concept $c$ created by $C$ in the inner loop, while $C$'s goal in the outer loop is to make $c$ \emph{unfamiliar} to $A$ at first but highly \emph{learnable} or \emph{memorable}.}
\label{fig:overview}
\end{figure}

Finally, an account of conceptual creativity must address aesthetics, i.e.\ what makes a creation appealing to our senses. Beyond evolutionary preferences, a powerful computational handle is abstraction through compression. Many artistic forms feature complex symmetries: Islamic geometric patterns, Gothic architecture, and the repetition and structure of musical composition are all highly compressible, in the sense that a small fragment suffices to predict the whole. Compressibility, however, is not universal: it depends on what the observer already knows. Aesthetic judgment is therefore partly an ``acquired taste''; Jazz can sound chaotic on first listen and rich once the listener has internalized its chord progressions and rhythmic motifs.

Together these examples sketch a picture of creativity at the \emph{frontier of understanding}. Existing concepts form a representation manifold built from past experience, and novel concepts lie just beyond it, past what an observer's current representations can immediately predict, but close enough that a small update can assimilate them (Figure~\ref{fig:intro}). \citet{schmidhuber2010formal} formalized this as \emph{compression progress}: a stimulus is interesting when it improves an observer's predictive model. As the title suggests, we hypothesize that \emph{an artifact is creative for an observer if it is initially unfamiliar to that observer and can be rapidly learned by it through a brief adaptation}. We operationalize unfamiliarity as the observer's initial prediction error $L_0$, and learnability as the drop from $L_0$ to the residual error $L_T$ after $T$ adaptation steps. The definition is observer-relative by construction, since the same artifact can be creative for one observer and familiar or unlearnable for another. We use adaptation, learning, and memorization interchangeably, because our focus is the fast adaptation of an observer to a single instance.

Motivated by this hypothesis, we propose a guided generation method through meta-learning. In the inner loop, the observer, which we call the \emph{Appraiser}, takes a few gradient steps to assimilate a candidate. In the outer loop, a \emph{Creator} refines the candidate so that it begins unfamiliar and ends memorable to that same observer (Figure~\ref{fig:creator_appraiser}). We instantiate the Creator as a frozen diffusion model, and the Appraiser as an autoencoder on MNIST and as CLIP with a low-rank text adapter on natural images. On both MNIST and natural images, this criterion alone guides the diffusion model to generate novel yet structured images that lie outside its training distribution.

\section{Creator-Appraiser Meta-Learning}

We define an \emph{observer} as a \emph{predictive system}: an encoder $E$ that maps an input $x$ to a representation $z = E(x)$, and a head $H$ that emits a prediction whose discrepancy from $x$ we write as $\mathcal{L}(H, E; x)$. The head can take many forms: a decoder that reconstructs $x$, a next-token predictor over a sequence, or a classifier that decides whether $z$ belongs to $x$ versus a set of reference stimuli. We say the observer \emph{understands} $x$ when the prediction error $\mathcal{L}(H, E; x)$ is small. This covers common self-supervised objectives such as reconstruction~\citep{vincent2008extracting,kingma2013auto}, next-token prediction~\citep{bengio2003neural}, masked prediction~\citep{devlin2019bert,he2022masked}, and latent prediction~\citep{lecun2022path,assran2023self}.

A Creator-relevant observer does not stay frozen. When faced with a novel stimulus that its current representations fail to predict, an adaptive observer updates a subset of its parameters to accommodate it, and the Creator gets useful feedback only because of this update. Concretely, novelty is the initial prediction error, and aesthetics is the rate at which a brief adaptation reduces that error. The value of a stimulus is then the size of this drop. We assign this adaptive observer the role of an \emph{Appraiser}: an observer that not only predicts but also reports its own learning progress back to the creator. We now formalize this in a bilevel optimization between a Creator that generates and an Appraiser that adapts.

\paragraph{The two-player formulation.}
We propose conceptual creation as a bilevel optimization between a Creator ($C$) that generates candidate concepts and an Appraiser ($A$) that adapts to them.

The Appraiser $A$ is a predictive system parameterized by a pair of weights $(\theta, \phi)$: $\theta$ are slow weights that hold $A$'s prior over its inputs, fixed throughout creation; $\phi$ are fast weights that adapt within the inner loop. Given a candidate $c$ from $C$, $A$ runs an inner-loop optimization to fit $c$, minimizing its prediction error on $c$ alone:
\begin{equation}
\min_\phi \mathcal{L}_A(\phi, \theta; c),
\end{equation}
where $\mathcal{L}_A$ is the predictive loss of $A$, instantiated as reconstruction error, classification error, or any other discrepancy a predictive head produces on $c$. The improvement in this loss after $T$ inner-loop steps, $\mathcal{L}_A(\phi_0, \theta; c) - \mathcal{L}_A(\phi_T, \theta; c)$, is the reward $A$ reports back to $C$. $C$'s objective is therefore to produce a $c$ for which $A$'s learning progress is large:
\begin{equation}
\min_c \mathcal{L}_C(c) = \min_c \mathcal{L}_A(\phi_T, \theta; c) - \mathcal{L}_A(\phi_0, \theta; c),
\end{equation}
where $\phi_0$ are the initial fast weights and $\phi_T$ are the weights after $T$ inner-loop steps; $C$ minimizes this composite objective with respect to its own parameters or sample space. 
The reward to $C$ is thus $A$'s learning progress, and a candidate earns a high reward when $A$ finds it unfamiliar at first and then learns it quickly. Many different candidates can earn a high reward, so the objective does not single out one artifact, and in our experiments different noise seeds lead the Creator to different outputs (Section~\ref{sec:natural_images}). Before each $C$ iteration, we reset the fast weights $\phi$ to a fixed initialization $\phi_0$, so $A$ meets every candidate as a first impression. Algorithm~\ref{alg:camel} summarizes the procedure, and Figure~\ref{fig:creator_appraiser} illustrates it.

\begin{figure}[t]
\centering
\includegraphics[width=\textwidth]{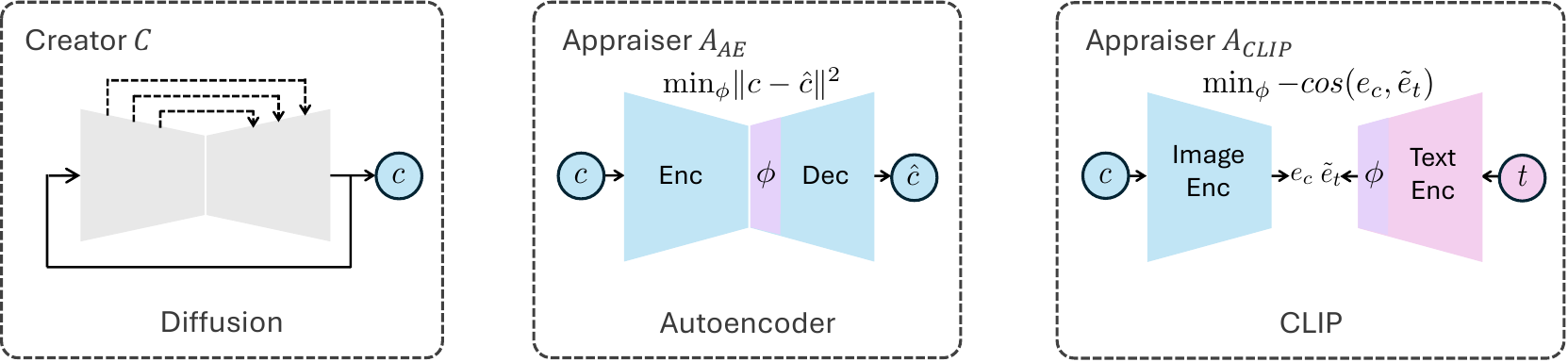}
\caption{The Creator-Appraiser framework instantiated with the Autoencoder and CLIP Appraisers. $\phi$ denotes the adaptation parameter in the Appraiser for concept learning.}
\label{fig:appraiser_instantiations}
\end{figure}

\begin{table}[t]
\centering
\caption{Appraiser implementation details.}
\label{tab:instantiations}
\small
\begin{tabular}{@{}lll@{}}
\toprule
 & Autoencoder Appraiser (MNIST) & CLIP Appraiser (natural images) \\
\midrule
Creator $C$ & DDPM, frozen & Stable Diffusion v1.5, frozen \\
Creator variable $\xi$ & noise latent & prompt text embedding \\
Slow weights $\theta$ & LeNet encoder, decoder & CLIP image and text encoders \\
Fast weights $\phi$ & first decoder layer & rank-$r$ LoRA on text projection \\
Inner loss $\mathcal{L}_A$ & pixel MSE & negative image-text cosine \\
Reward $R$ & $(L_0 - L_T)$ with gates $g_0, g_T$ & $\text{sim}_T - \text{sim}_0$ \\
Constraint on $R$ & reward shaping ($\mu$, $\sigma$, $\tau$) & adapter rank $r$ \\
Evaluation & $(L_0, L_T)$ plane, reward ablation & DINOv2 reachability, Qwen2.5-VL relevance \\
\bottomrule
\end{tabular}
\end{table}

\begin{algorithm}[h]
\caption{Creator-Appraiser Meta-Learning (one outer iteration)}
\label{alg:camel}
\begin{algorithmic}[1]
\Require Creator $C$ with learnable variable $\xi$; Appraiser $A$ with slow weights $\theta$ and fast-weight initialization $\phi_0$; loss $\mathcal{L}_A$; inner steps $T$, inner LR $\alpha$; outer LR (guidance scale) $\eta$.
\State $\tilde{c} \gets C(\xi)$ \Comment{generate (intermediate or final) candidate}
\State $\phi \gets \phi_0$ \Comment{reset Appraiser fast weights for first impression}
\State $L_0 \gets \mathcal{L}_A(\phi, \theta;\, \tilde{c})$
\For{$k = 1, \ldots, T$} \Comment{inner-loop adaptation, differentiable in $\xi$}
    \State $\phi \gets \phi - \alpha\, \nabla_\phi \mathcal{L}_A(\phi, \theta;\, \tilde{c})$
\EndFor
\State $L_T \gets \mathcal{L}_A(\phi, \theta;\, \tilde{c})$
\State $R \gets L_0 - L_T$ \Comment{learning-progress reward}
\State $\xi \gets \xi + \eta\, \nabla_\xi R$ \Comment{meta-gradient update on Creator}
\end{algorithmic}
\end{algorithm}

In our experiments, the outer iteration is run once per denoising step of a diffusion Creator, with $\tilde{c}$ being the predicted clean output $\hat{x}_0^{(t)}$ at step $t$ and $\xi$ being either the noise latent (MNIST) or the text embedding (natural images). The same template applies to any iterative or one-shot generator with a differentiable parameter $\xi$.

\subsection{Appraiser instantiations}
\label{sec:appraiser_instantiations}

The framework accepts any adaptive predictive model as the Appraiser. Because creativity is defined relative to an observer, the Appraiser's prior and loss decide which artifacts count as creative. A reconstruction loss rewards strokes and textures the decoder can reproduce, and a classifier loss rewards movement across fixed class boundaries. A loss in a joint image-text space rewards images that a small adaptation can align with the prompt, which favors reinterpretations of the concept. We therefore match the Appraiser to the domain. We use an Autoencoder Appraiser for MNIST and a CLIP Appraiser for natural images (Figure~\ref{fig:appraiser_instantiations} and Table~\ref{tab:instantiations}).

\subsubsection{Autoencoder Appraiser}
\label{sec:autoencoder_appraiser}

The encoder $E$ is frozen and pretrained for classification, and the decoder $H$ is a lightweight network pretrained to reconstruct training images from $E$'s features. The inner loop fine-tunes the decoder on the single candidate image $c$ with reconstruction loss (MSE), and the reward is the loss improvement after $T$ steps. The Appraiser thus rewards inputs the decoder can quickly assimilate from its existing representations, and gives near-zero reward on noise-like inputs that the decoder cannot fit at all.

\paragraph{Reward shaping.} We refer to $\mathcal{L}_{A,0}$ as the \emph{unfamiliarity} of $c$ (how surprising it is to a naive observer) and $\mathcal{L}_{A,T}$ as the \emph{residual unfamiliarity} (how much remains unpredictable after adaptation). For the autoencoder Appraiser, the raw improvement $\mathcal{L}_{A,0} - \mathcal{L}_{A,T}$ over-rewards high-noise inputs (large initial reconstruction loss to recover from) even when no meaningful structure has emerged. We therefore multiply the improvement by a Gaussian complexity gate and an exponential success gate, which constrain $\mathcal{L}_{A,0}$ and $\mathcal{L}_{A,T}$ to a ``learnable yet unfamiliar'' region:
\begin{equation}
    R = (\mathcal{L}_{A,0} - \mathcal{L}_{A,T}) \cdot g_0(\mathcal{L}_{A,0}) \cdot g_T(\mathcal{L}_{A,T}),
\end{equation}
\begin{equation}
    g_0(\mathcal{L}_{A,0}) = \exp\!\left(-\frac{(\mathcal{L}_{A,0} - \mu)^2}{2\sigma^2}\right), \qquad
    g_T(\mathcal{L}_{A,T}) = \exp\!\left(-\frac{\mathcal{L}_{A,T}}{\tau}\right).
\end{equation}
The \emph{complexity gate} $g_0$ keeps the initial prediction error in a target range around $\mu$ (avoiding both familiar low-$\mathcal{L}_{A,0}$ inputs and noise-like high-$\mathcal{L}_{A,0}$ inputs). The \emph{success gate} $g_T$ rewards low final error, ensuring the decoder actually fits the input. We show the effect of applying the gates in Figure~\ref{fig:reward_ablation}.

\subsubsection{CLIP Appraiser}
\label{sec:clip_appraiser}

The Autoencoder Appraiser ties learnability to a domain-specific encoder--decoder pair pretrained on the target distribution. This is a good fit for settings like MNIST, where images are small and a single-modality reconstruction loss is sufficient. Natural image generation is different in two ways. First, many pretrained creative models are \emph{text-conditioned}: the prompt names the concept and the Appraiser must judge whether the candidate fits that concept, so the learnability signal needs to be present in a joint vision-language space. Second, pretraining a high-fidelity natural-image decoder is expensive, and the reconstruction loss would be dominated by pixel-level texture rather than concept-level semantics. Consistent with this, in preliminary experiments, DINOv2-based reconstruction and classifier Appraisers produced degenerate images, likely because the reconstruction objective is low-level and the classifier lacks constraints and neighboring concepts to contrast with. We therefore instantiate the Appraiser on top of a frozen vision-language encoder pair, where the inner loop adapts a small text-side projection to match the generated image, providing a concept-level learnability signal that is naturally text-conditioned.

The CLIP Appraiser measures how quickly a small text-side update can learn to match a generated image (Figure~\ref{fig:appraiser_instantiations}). We use the contrastive image-text model CLIP~\citep{radford2021learning} with both encoders frozen. The inner loop adapts the text projection layer with a low-rank adapter~\citep{hu2022lora} of rank $r$, $W_{\text{text}}' = W_{\text{text}} + UV$, with $U$ initialized to zero. Over $T$ steps, it updates $U$ and $V$ to maximize $\cos(e_{\text{img}}, \tilde{e}_{\text{text}})$, where $\tilde{e}_{\text{text}}$ is the adapted text feature. The reward is the gain in image-text similarity, $R = \text{sim}_T - \text{sim}_0$.

The rank $r$ limits the capacity of the adaptation and takes the place of the Gaussian gates of the Autoencoder Appraiser. At small $r$, the inner loop cannot fit an arbitrary image direction in $T$ steps, so the meta-gradient stays near the prompt's concept. A larger $r$ relaxes this limit and lets the outputs move farther from the prompt (Section~\ref{sec:natural_images}). With this limit in place, we use the similarity gain as the reward without further shaping.


\section{Experiments}

We evaluate our framework in two settings: (1)~an MNIST setting that validates the core learnability signal and the unfamiliarity--learnability frontier with the Autoencoder Appraiser, and (2)~a natural image setting using Stable Diffusion with the CLIP Appraiser.

\subsection{MNIST}

\paragraph{Setup.} The Appraiser uses a LeNet-style encoder (Conv$\to$ReLU$\to$Pool, two layers, 512-dimensional feature space) and a transposed-convolutional decoder that reconstructs 32$\times$32 grayscale images from the feature bottleneck. Both are pretrained on MNIST: the encoder with classification loss, the decoder with pixel-level MSE reconstruction loss. The Creator is a pretrained DDPM diffusion model~\citep{ho2020denoising} that generates MNIST-like images. For brevity in figures and text we write $L_0 = \mathcal{L}_{A,0}$ and $L_T = \mathcal{L}_{A,T}$.

\begin{figure}[t]
\centering
\begin{subfigure}[c]{0.51\textwidth}
  \centering
  \includegraphics[width=\textwidth]{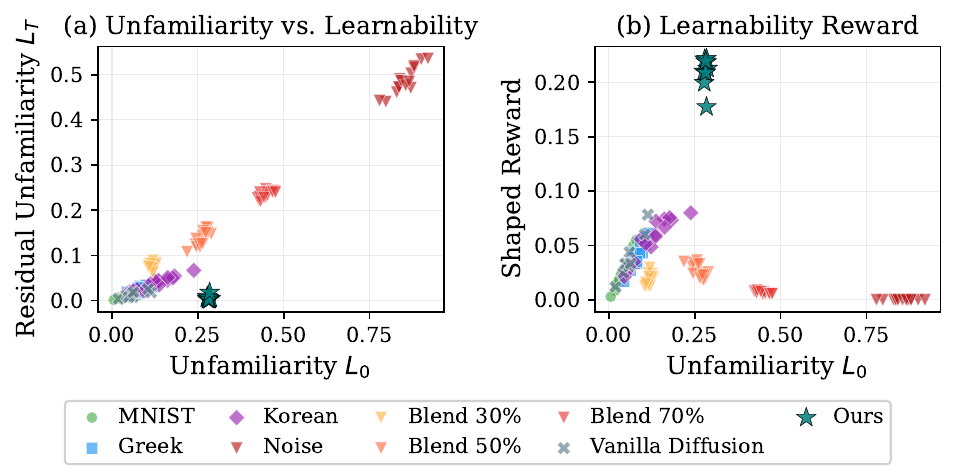}
  \caption{Unfamiliarity--learnability space.}
  \label{fig:learnability_space}
\end{subfigure}
\hfill
\begin{subfigure}[c]{0.47\textwidth}
  \centering
  \includegraphics[width=\textwidth]{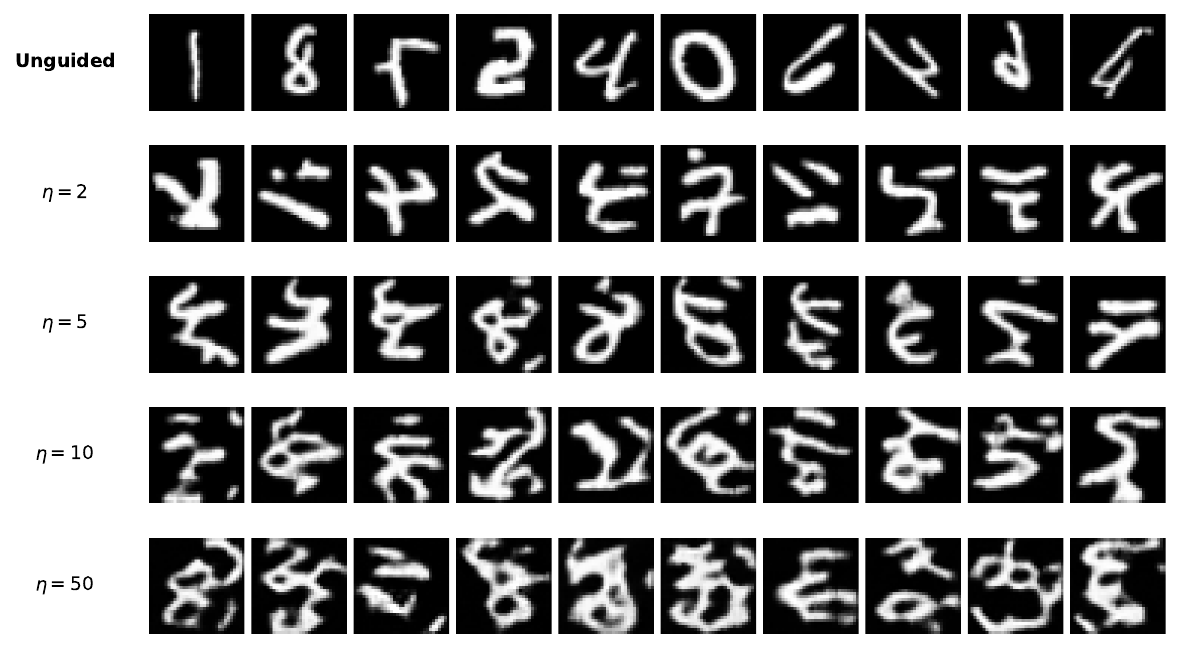}
  \caption{Guided samples at increasing $\eta$.}
  \label{fig:generated_grid}
\end{subfigure}
\caption{\textbf{(a)} Guided images are moderately unfamiliar yet quickly learnable, and their shaped reward exceeds that of the best reference category (Korean). \textbf{(b)} Larger $\eta$ gives more novel and complex samples that keep a stroke-like structure. At the largest $\eta$, they no longer match any digit.}
\label{fig:mnist_main}
\end{figure}

\paragraph{Learnability signal validation.} In Figure~\ref{fig:learnability_space}, we probe whether the shaped reward correctly orders stimuli that span the unfamiliarity--learnability space, comparing MNIST test digits (familiar, structured), Greek and Korean characters from Omniglot (unfamiliar, structured), Gaussian noise (unfamiliar, unstructured), and noise-blended MNIST digits at 30\%, 50\%, and 70\% noise ratios. The blends have similar $L_0$ to the Omniglot characters but lack coherent structure. In the $(L_0, L_T)$ plane (left panel), noise lies in the top-right corner, unfamiliar and unlearnable. Familiar digits lie in the bottom-left, and our guided generations lie in the bottom-middle, moderately unfamiliar yet quickly learnable. The shaped reward (right panel) ranks Korean and Greek above noise blends with the same $L_0$ and gives noise a near-zero reward.

\begin{figure}[t]
\centering
\includegraphics[width=\textwidth]{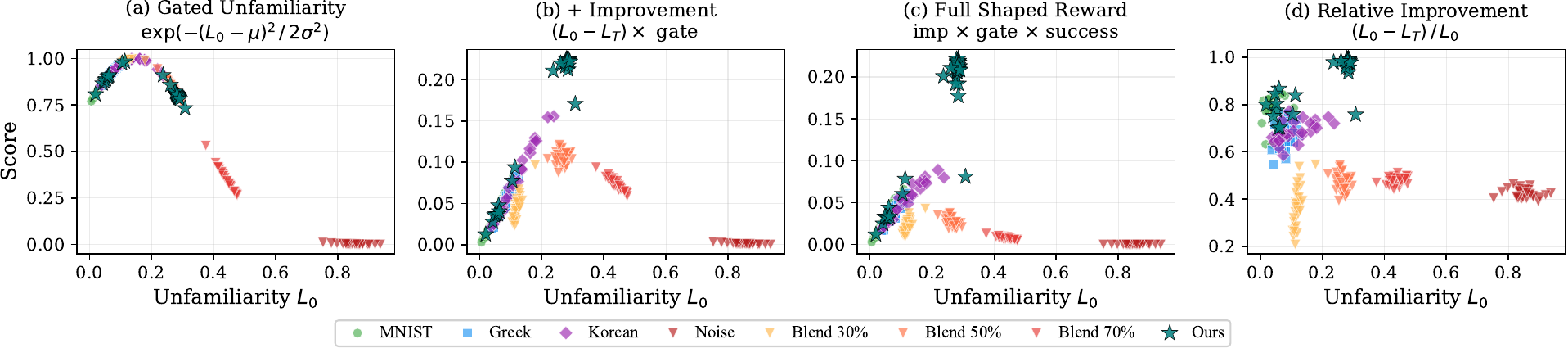}
\caption{Decomposition of the shaped reward. \textbf{(a)}~The complexity gate alone cannot distinguish structured from degraded stimuli at similar $L_0$. \textbf{(b)}~Adding the improvement term separates structured scripts from noise blends. \textbf{(c)}~The full shaped reward. \textbf{(d)}~Relative improvement inverts the desired ranking, rewarding familiarity over structured novelty.}
\label{fig:reward_ablation}
\end{figure}

\paragraph{Hypothesis test.} This experiment tests the central claim of the paper, that the improvement term $L_0 - L_T$ adds signal beyond unfamiliarity $L_0$. The null hypothesis is that the shaped reward reduces to gated unfamiliarity. Figure~\ref{fig:reward_ablation} ablates each reward component. In Figure~\ref{fig:reward_ablation}a, the complexity gate alone cannot separate Korean characters from noise-blended digits at the same $L_0$, because both pass the gate. In Figure~\ref{fig:reward_ablation}b, adding the improvement term separates them, since the decoder reduces its loss on Korean characters much more than on noise blends with comparable initial error. In the overlap zone $L_0 \in [0.05, 0.15]$, the rank correlation with the shaped reward is $\rho = -0.10$ for the gate and $\rho = 0.89$ for the improvement term, which contradicts the null hypothesis. In Figure~\ref{fig:reward_ablation}c, the full reward adds the success gate $g_T$ to suppress inputs the decoder cannot fit. In Figure~\ref{fig:reward_ablation}d, the \emph{relative} improvement $(L_0 - L_T)/L_0$ reverses the ranking and rewards familiar inputs over structured novel ones.

\paragraph{Layer-specific learnability.} We further ablate which decoder layers are fine-tuned during the inner loop (Appendix Figure~\ref{fig:layer_ablation}). Fine-tuning only the \emph{first} (high-level, semantic) decoder layer yields improvement comparable to fine-tuning the full decoder, while fine-tuning only the \emph{last} (low-level, textural) layer yields near-zero improvement, so learnability is driven by adaptation of semantic features.

\paragraph{Guidance scale and ideal initial loss.} We vary the guidance scale $\eta$ and the complexity-gate target $\mu$ (Appendix Figures~\ref{fig:scale_analysis} and~\ref{fig:mu_analysis}). A larger $\eta$ gives more learnable but visually busier samples. Raising $\mu$ increases $L_0$ while $L_T$ stays nearly constant, so the gate sets the level of unfamiliarity without imposing a hard threshold.

\begin{figure}[t]
\centering
\includegraphics[width=\textwidth]{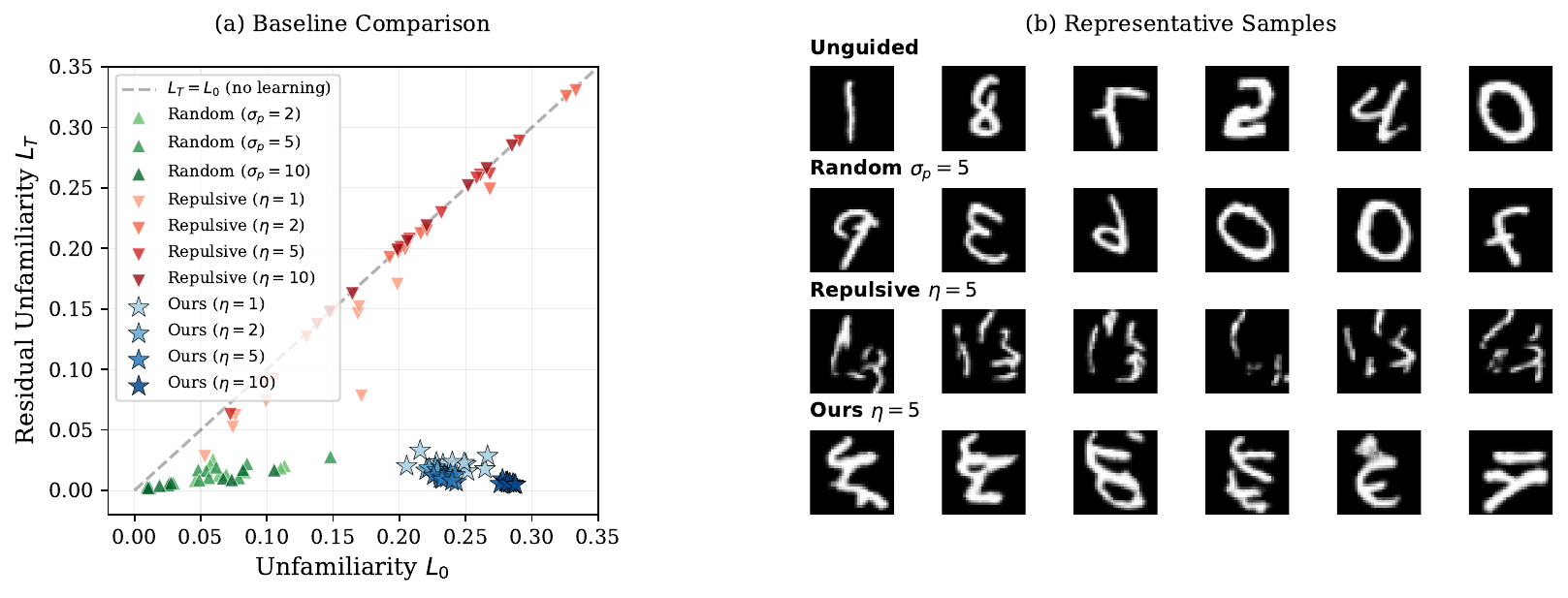}
\caption{Baseline comparison. \textbf{(a)}~Each point is one generated image in the $(L_0, L_T)$ plane; the dashed diagonal marks $L_T = L_0$ (no learning). Color gradients show increasing $\eta$ (or $\sigma_p$) within each method. In the legend, Random is noise perturbation and Unguided is vanilla DDPM. Noise perturbation stays familiar (near origin). Repulsive guidance increases unfamiliarity but rides the diagonal where no learning occurs. Only our method achieves the bottom-right: unfamiliar \emph{and} learnable. \textbf{(b)}~Representative samples at $\eta = 5$ ($\sigma_p = 5$ for noise perturbation).}
\label{fig:baseline_comparison}
\end{figure}

\paragraph{Comparison with baselines.} We compare against three baselines. \textbf{Vanilla DDPM} samples without guidance. \textbf{Noise perturbation} adds noise of scale $\sigma_p \in \{2, 5, 10, 20\}$ to the latent at each step. 
\textbf{Repulsive guidance}, a variant inspired by ProCreate~\citep{lu2024procreate}, pushes the encoder features of the predicted clean image away from the digit class prototypes with scale $\eta \in \{1, 2, 5, 10, 20, 50, 100\}$, which gives deviation without learnability. In Figure~\ref{fig:baseline_comparison}a, noise perturbation stays near the origin because the DDPM absorbs the perturbation. Repulsive guidance follows the no-learning diagonal $L_T \approx L_0$, and only our method reaches the bottom-right. Optimizing unfamiliarity alone thus tends to push samples in directions the Appraiser cannot learn. In Figure~\ref{fig:baseline_comparison}b, noise-perturbed samples look like unguided digits, repulsive-guided samples are disrupted but unstructured, and ours are coherent stroke-like glyphs. Additional samples from each method are in Appendix Figure~\ref{fig:baseline_grid}.

\subsection{Natural Images}
\label{sec:natural_images}

\begin{figure}[p]
\centering
\includegraphics[width=0.9\textwidth]{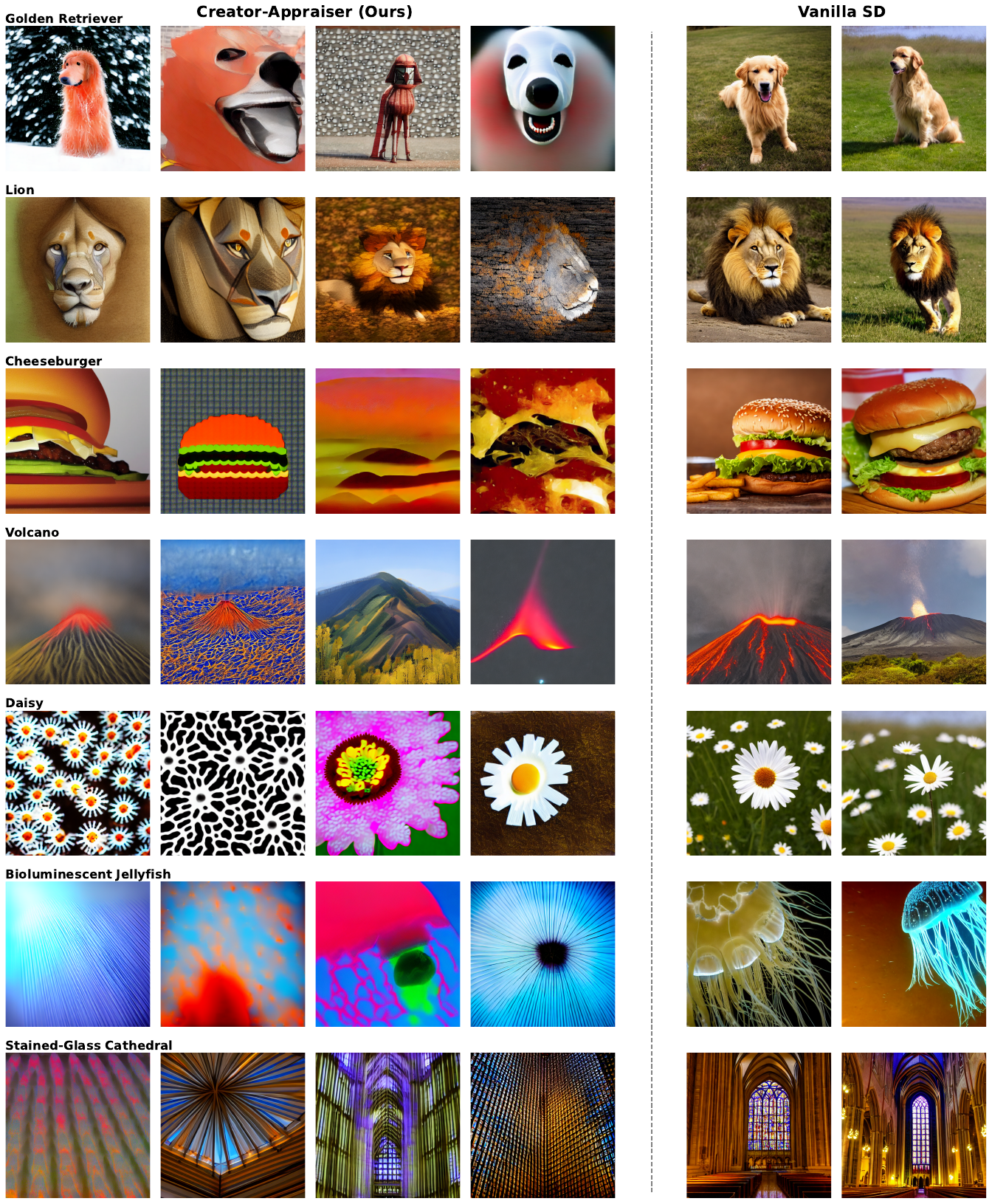}
\caption{Curated outputs from the CLIP Appraiser at LoRA rank $r \in \{8, 32\}$, $\eta=10$. Each row shows four samples from our method (left) and two vanilla SD samples (right) for the same prompt; rows 1--5 are ImageNet class-conditioned, rows 6--7 are open-ended text prompts. The meta-gradient produces visually distinct stylistic interpretations within each prompt.}
\label{fig:clip_lora_picks}
\end{figure}

\paragraph{Setup.} We instantiate the framework on natural image generation using the CLIP Appraiser (Section~\ref{sec:clip_appraiser}) paired with a Stable Diffusion v1.5 Creator~\citep{rombach2022high}. Unlike the MNIST setting, where the Creator's learnable variable is the latent noise $z$ and the meta-gradient flows back to $z$ at each denoising step, here the learnable variable is the \emph{text embedding}: the prompt is encoded once, the encoded representation becomes a learnable parameter, and the meta-gradient updates it at each denoising step. We found this search in semantic space necessary for the reward to track semantics rather than pixel texture. We evaluate two types of prompts: \textbf{(i) class-conditioned prompts} of the form ``a photo of a \{class\}'' for ImageNet classes, and \textbf{(ii) open-ended prompts} drawn from a curated set of stylistic descriptions (e.g., ``stained-glass cathedral interior at night''). For each prompt we sweep LoRA rank $r \in \{4, 8, 16, 32\}$ at outer-loop guidance $\eta = 10$.

\paragraph{Measures.} We report two quantities from the CLIP Appraiser and two measures from separate evaluator models. The Appraiser quantities are (i) \textbf{$\text{sim}_0$}, the cosine similarity between the generated image's CLIP embedding and the prompt's text embedding under the \emph{frozen} CLIP model, which is also the inner-loop initial similarity, and (ii) \textbf{$\Delta\text{sim} = \text{sim}_T - \text{sim}_0$}, the similarity improvement after the $T$-step LoRA inner loop, which is the reward the meta-gradient optimizes. The evaluators do not take part in generation. (iii) The \textbf{relevance rate} is the fraction of samples that Qwen2.5-VL-7B~\citep{bai2025qwen2vl}, a vision-language model (VLM), scores at least $5$ on a $0$--$10$ scale of concept presence (Appendix~\ref{sec:appendix_bon}). (iv) The \textbf{DINOv2 distance} is the nearest-neighbor distance from a sample to 1000 vanilla samples of the same prompt in DINOv2 feature space~\citep{oquab2024dinov2}, described below.

\begin{figure}[t]
\centering
\includegraphics[width=0.55\textwidth]{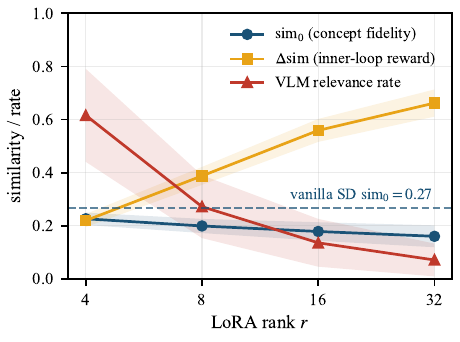}
\caption{Effect of LoRA rank. Lines show the mean $\pm 1$ s.d.\ across 20 prompts. As $r$ grows, the inner-loop reward $\Delta\text{sim}$ rises, frozen-CLIP $\text{sim}_0$ drifts below the vanilla level, and the Qwen2.5-VL relevance rate falls. Per-prompt values in Table~\ref{tab:natural_image_quant_full}.}
\label{fig:rank_tradeoff}
\end{figure}

\paragraph{Results.} Figure~\ref{fig:clip_lora_picks} shows curated outputs from our method next to vanilla SD samples for the same prompts. The vanilla samples are photorealistic and on-prompt. By contrast, our outputs include forms that the prompt does not describe: a lion in autumn-leaf form, golden retrievers in novel coloration and proportion, a cheeseburger rendered as a sunset landscape, a daisy whose petals dissolve into ink strokes, and a pastel oil-painted volcano. The input string was simply ``a photo of a daisy'' or ``a photo of a lion'', and the diffusion weights are not fine-tuned. The deviation is found inside the prompt embedding by the meta-gradient on the inner-loop learning-progress signal.

\paragraph{Capacity-vs-fidelity trade-off.} As rank grows (Figure~\ref{fig:rank_tradeoff}), $\Delta\text{sim}$ rises while $\text{sim}_0$ and the relevance rate fall. A low-rank inner LoRA cannot fit off-manifold images, which keeps the meta-gradient near the prompt's class; higher rank lifts that constraint.

\begin{figure}[t]
\centering
\includegraphics[width=\textwidth]{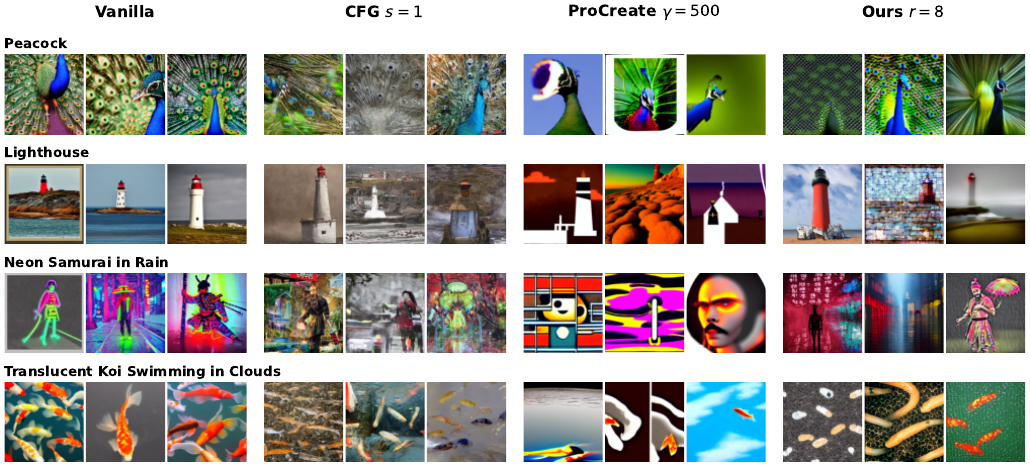}
\caption{Baselines and ours ($r{=}8$) on four prompts. Samples are random draws with Qwen2.5-VL relevance $\geq 5$ (vanilla ungated). 
}
\label{fig:baseline_compare}
\end{figure}

\paragraph{Reachability beyond vanilla.} We ask whether our outputs are samples that vanilla Stable Diffusion would also produce given enough draws. For each of the 20 prompts we draw $N=1000$ vanilla samples, embed them and our samples with a frozen DINOv2 ViT-L/14~\citep{oquab2024dinov2}, and compute nearest-neighbor cosine distance to the vanilla set. We keep only our samples with Qwen2.5-VL relevance $\geq 5$. At $r{=}8$, our samples lie a median $4.6\times$ farther from the vanilla set than vanilla samples lie from each other, and this distance grows with LoRA rank. The meta-gradient therefore reaches outputs farther from the vanilla set than vanilla sampling does at $N=1000$. Table~\ref{tab:baselines} compares baselines on the same prompts, with samples in Figure~\ref{fig:baseline_compare}. In those samples, CFG $s{=}1$ outputs are desaturated and blurry, ProCreate $\gamma{=}500$ outputs often become flat graphic shapes, and ours vary in color, texture, and composition. Baselines that lower the classifier-free guidance (CFG) scale or repel samples from reference images reach a similar trade-off between distance and relevance rate. Our method reaches it through the learnability objective alone, without an explicit deviation term. The increase with rank also holds without the relevance gate and at a fixed sample count per prompt, and Appendix~\ref{sec:appendix_reachability} gives per-prompt numbers and nearest-neighbor pairs. Appendix~\ref{sec:appendix_bon} shows uncurated top-$K$ samples.

\begin{table}[t]
\centering
\caption{Baselines on 20 prompts. Rate: fraction with Qwen2.5-VL relevance $\geq 5$. NN dist.: median DINOv2 distance of those samples to the 1000 vanilla samples. Vanilla distance is to the other 999 samples. 
Full results in Appendix Table~\ref{tab:baselines_full}.
}
\label{tab:baselines}
\small
\begin{tabular}{@{}lcc@{\hspace{2em}}lcc@{}}
\toprule
Baseline & Rate & NN dist. & Ours & Rate & NN dist. \\
\midrule
Vanilla (CFG 7.5) & 0.96 & 0.11 & $r = 4$ & 0.62 & 0.41 \\
CFG $s \in \{3, 5, 10\}$ & 0.94--0.97 & 0.06--0.17 & $r = 8$ & 0.27 & 0.53 \\
CFG $s = 1$ & 0.64 & 0.47 & $r = 16$ & 0.14 & 0.55 \\
CFG $s = 0.5$ & 0.17 & 0.66 & $r = 32$ & 0.07 & 0.66 \\
ProCreate~\citep{lu2024procreate}, $\gamma = 250$ & 0.73 & 0.38 & & & \\
ProCreate, $\gamma = 500$ & 0.28 & 0.58 & & & \\
ProCreate, $\gamma = 1000$ & 0.08 & 0.72 & & & \\
\bottomrule
\end{tabular}
\end{table}

\section{Related Work}\label{sec:related}

\paragraph{Computational creativity.}
A persistent question is whether machines can create at all. Ada Lovelace famously argued that Babbage's Analytical Engine ``has no pretensions whatever to \emph{originate} anything'' \citep{lovelace1843notes}. We take the position that creativity is best defined operationally, by its \emph{effect on an observer} rather than by the creator's experiential phenomenology~\citep{nanay2014experiential}. On that definition, machines can create, and the question becomes how to build them. The broader field of Computational Creativity has modeled creative behavior in machines for decades (see survey~\citep{franceschelli2024creativity}). \citet{boden2004creative} distinguishes \emph{exploratory} creativity, which works within an existing conceptual space, from \emph{transformational} creativity, which reshapes the space itself. We do not assign outputs to one of these categories. In our experiments the Creator is frozen, so its outputs explore an existing space, and the Appraiser learns each output in its inner loop.

\citet{schmidhuber2010formal} proposed a formal theory of creativity based on compression progress: a stimulus is ``interesting'' to an observer when it improves the observer's predictive model. The key difference in our work is that the Creator is \emph{aware} of the Appraiser's inner-loop learning progress: it is updated by meta-gradient through the Appraiser's adaptation, not just by a scalar reward. Realizing this in a modern vision generative model requires differentiating through diffusion sampling and through an inner loop on a CLIP-LoRA Appraiser. Follow-up work (e.g.\ POWERPLAY~\citep{srivastava2013powerplay}) aimed at RL task completion rather than visual creation. Our reward shaping also resonates with \citet{berlyne1960conflict}'s arousal theory in psychology, where aesthetic pleasure peaks at intermediate novelty and complexity, neither too simple nor too overwhelming.

\paragraph{Information theory and bounded learnability.}
\emph{Bayesian surprise}~\citep{itti2009bayesian} measures how much an observation moves an observer, as the KL divergence between posterior and prior beliefs under a fixed probabilistic model. Our reward instead measures the loss reduction from a bounded gradient update, and the Appraiser's loss plays the role of the likelihood. Classical information measures abstract away the bounded observer: Shannon entropy treats pure noise as maximally informative, while Kolmogorov complexity~\citep{kolmogorov1965three} assumes an unbounded observer for whom any compressible structure is free. Neither captures structure a \emph{bounded} learner can extract with a small adaptation. 
Most closely related, \citet{finzi2026epiplexity} recently introduced \emph{epiplexity}: the length of the shortest program achieving the optimal two-part code at a given compute budget. Our Creator-Appraiser reward $\mathcal{L}_{A,0} - \mathcal{L}_{A,T}$ measures observer-relative bounded learnability via a \emph{specific} Appraiser rather than a search over learners. Crucially, epiplexity is a measurement on existing data, whereas our framework adds a generative side, with the Creator producing new artifacts by optimizing the Appraiser's reward via meta-learning gradients.

\paragraph{Deep generative models and creativity.}
GANs~\citep{goodfellow2014generative,karras2019style} and diffusion models~\citep{ho2020denoising} are the dominant paradigm for high-fidelity image generation. Creative Adversarial Networks~\citep{elgammal2017can} extended GANs by training the generator to deviate from learned artistic styles while remaining aesthetically plausible. 
However, the apparent creativity of these models is, by default, largely a consequence of recombining concepts and styles already present in the training distribution~\citep{richardson2024conceptlab,feng2025distribution}, and obtaining outputs further from typical sampling generally requires careful prompt engineering.

Several recent works have proposed explicit mechanisms for creative generation in diffusion models. \citet{lu2024procreate} guide diffusion away from existing training examples using a propulsive energy term that repels generated samples from specific reference images. \citet{han2025enhancing} enhance creativity by amplifying low-frequency components in early diffusion layers. \citet{golan2025vlm} use vision-language models to adaptively construct negative prompts that steer generation toward unconventional outputs. These methods share a strategy of pushing the generator away from its typical outputs. We instead optimize for \emph{learnability}: the output must be unfamiliar to the observer but also structured enough to be learned quickly.

\paragraph{Adversarial, actor-critic, and meta-learning frameworks.}
The Creator-Appraiser structure resembles the generator-discriminator dynamic in GANs~\citep{goodfellow2014generative} and the actor-critic framework in RL~\citep{sutton2018reinforcement}, with key differences. Unlike a GAN discriminator, which trains toward a fixed decision boundary at equilibrium, the Appraiser is a \emph{dynamic learner} re-initialized for each creation, and the Creator is rewarded not for fooling it but for the inner-loop \emph{learning progress} on that artifact. The actor-critic analogy is closer: like a critic the Appraiser scores artifacts, but its ``value'' is the \emph{learning progress} the Appraiser itself makes when adapting to the just-produced artifact, not a long-horizon return, in the spirit of intrinsic-motivation rewards based on prediction improvement~\citep{schmidhuber2010formal,pathak2017curiosity}. The inner-loop adaptation draws on meta-learning~\citep{finn2017model}, and the text-embedding optimization relates to Textual Inversion~\citep{gal2023textual}, though with a learnability rather than reconstruction objective.


\section{Conclusion}
\label{sec:conclusion}
We started from the view that an artifact is creative for an observer when it lies just beyond what the observer can predict but remains learnable. Our Creator-Appraiser framework turns this view into a meta-learning objective, since the Creator can follow the Appraiser's learning progress through the inner loop. We tested the framework with MNIST digits and natural images. On MNIST, learning progress separates structured from unstructured stimuli, and in both settings its meta-gradient moves a frozen diffusion model away from its typical outputs. As future work, an Appraiser with memory across creations could judge an artifact learnable in light of what it has recently seen, which opens room for context-dependent and curriculum-driven creativity. The Creator-Appraiser structure also applies to any generator paired with an adaptive predictor, and extending to more modalities is the natural next step.

\vspace{0.15in}
\paragraph{Limitations.} Our observers are computational: an autoencoder and a CLIP Appraiser, with DINOv2 and Qwen2.5-VL as evaluators. We have not tested whether people judge the outputs as creative. 
The CLIP Appraiser also does not enforce fidelity to the prompt, and the relevance rate falls as the rank grows (Table~\ref{tab:baselines}). A relevance term in the reward could keep more samples on-prompt. Finally, our framing covers perceptual value only, rather than functional value. 

\section*{Acknowledgment}
This work was supported in part by Visko Platform, Pionex, a Google TPU Award, and the Institute of Information \& Communications Technology Planning Evaluation (IITP) under grants RS-2024-00469482 and RS-2024-00509279, funded by the Ministry of Science and ICT (MSIT) of the Republic of Korea in connection with the Global AI Frontier Lab International Collaborative Research. The compute is supported by the NYU High Performance Computing resources, services, and staff expertise.

\bibliographystyle{apalike}
\bibliography{ref}

\clearpage
\appendix
\section*{Appendix}
\addcontentsline{toc}{section}{Appendix}
\renewcommand{\thesection}{\Alph{section}}
\setcounter{section}{0}

\section{Additional MNIST Results}

\begin{figure}[h]
\centering
\includegraphics[width=\textwidth]{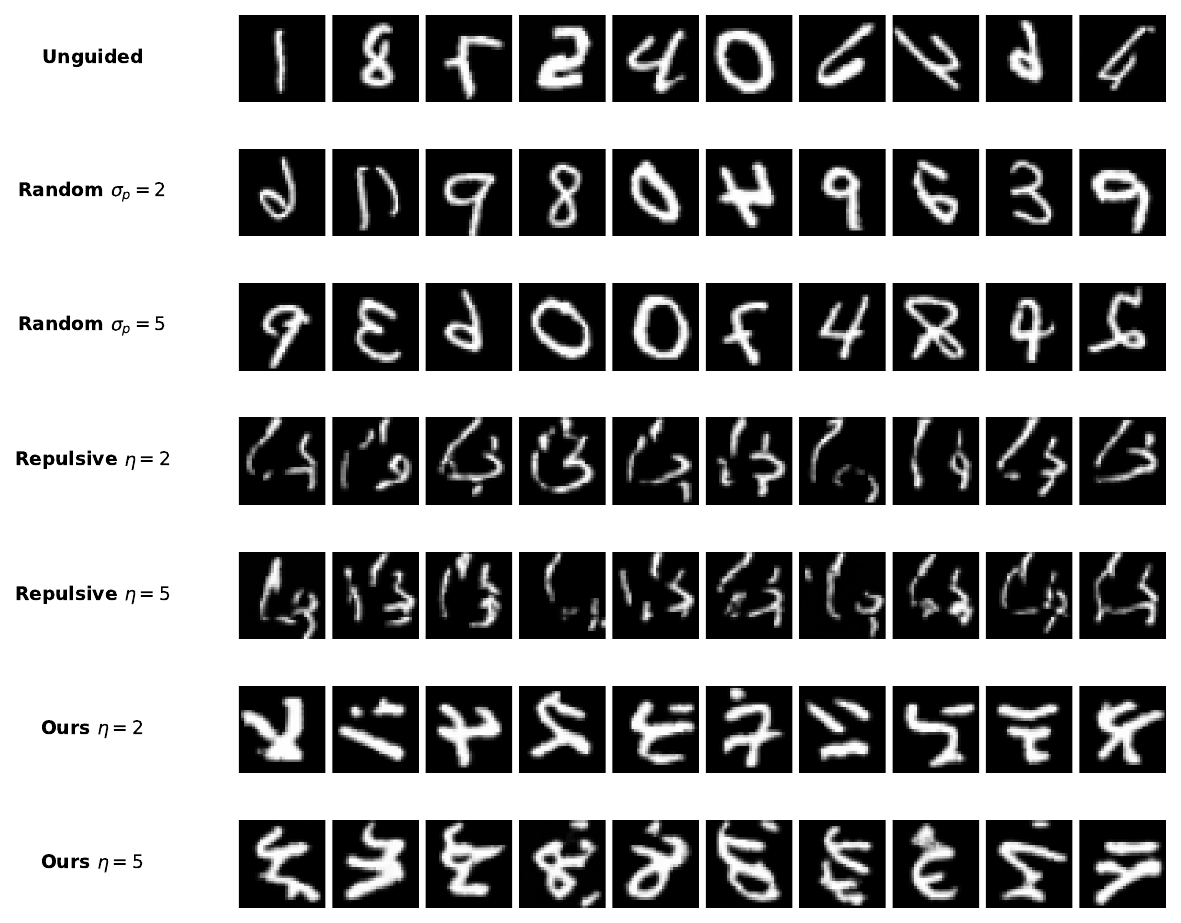}
\caption{Representative images from each method. \textbf{Unguided} (vanilla DDPM): standard DDPM digits. \textbf{Random} (noise perturbation, $\sigma_p = 2, 5$): outputs look like unguided samples, because the diffusion process absorbs the noise. \textbf{Repulsive guidance} ($\eta = 2, 5, 10$): images become more corrupted, and higher repulsion gives more unfamiliar and less coherent outputs. \textbf{Ours} ($\eta = 2, 5$; $\mu = 0.08$): novel, glyph-like patterns with coherent stroke structure.}
\label{fig:baseline_grid}
\end{figure}

\begin{figure}[h]
\centering
\includegraphics[width=0.62\textwidth]{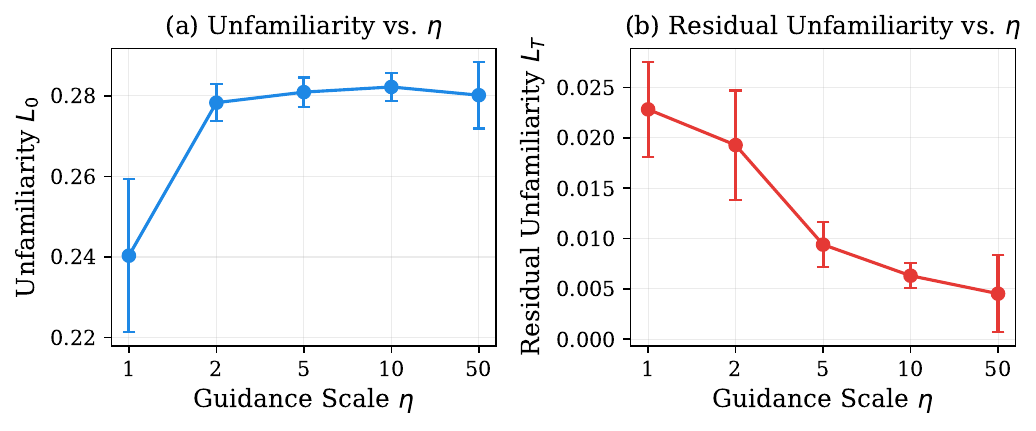}
\caption{Effect of guidance scale $\eta$ on generated image statistics ($\mu = 0.15$). \textbf{(a)}~Unfamiliarity $L_0$ saturates quickly beyond $\eta = 2$. \textbf{(b)}~Residual unfamiliarity $L_T$ falls as $\eta$ grows, so outputs become more learnable.}
\label{fig:scale_analysis}
\end{figure}

\begin{figure}[h]
\centering
\includegraphics[width=\textwidth]{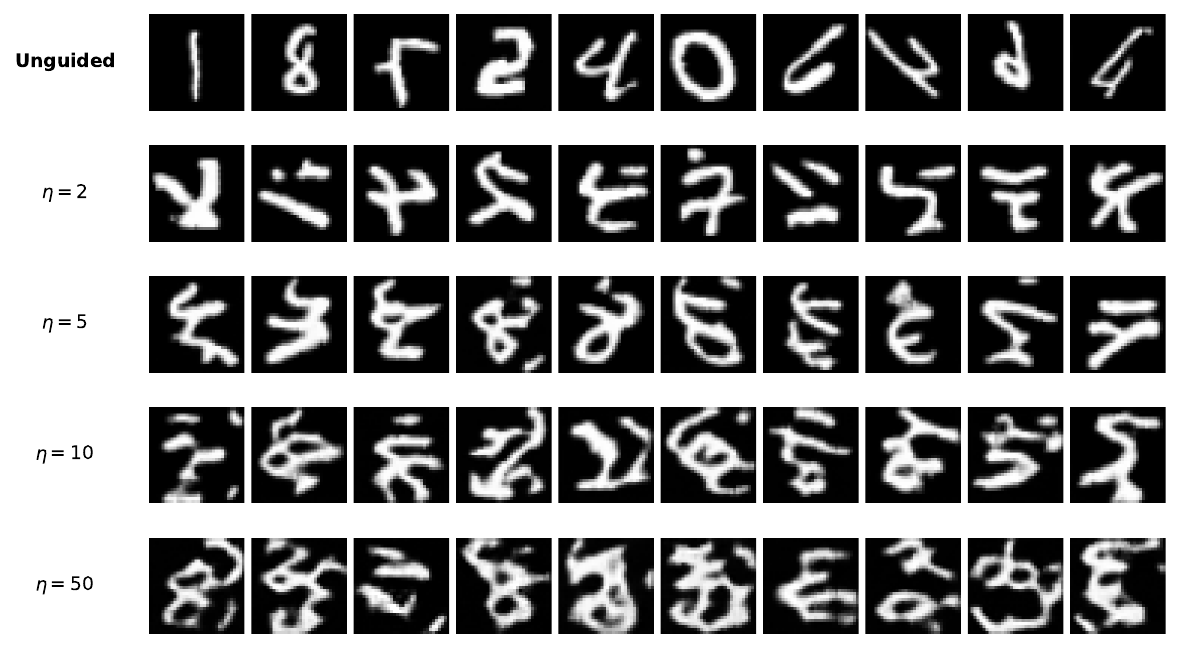}
\caption{Generated images at increasing guidance scales $\eta$, corresponding to the analysis in Figure~\ref{fig:scale_analysis}. Vanilla DDPM samples resemble MNIST digits. Guided outputs become more novel and keep a stroke-like structure.}
\label{fig:scale_grid_appendix}
\end{figure}

\begin{figure}[h]
\centering
\includegraphics[width=0.62\textwidth]{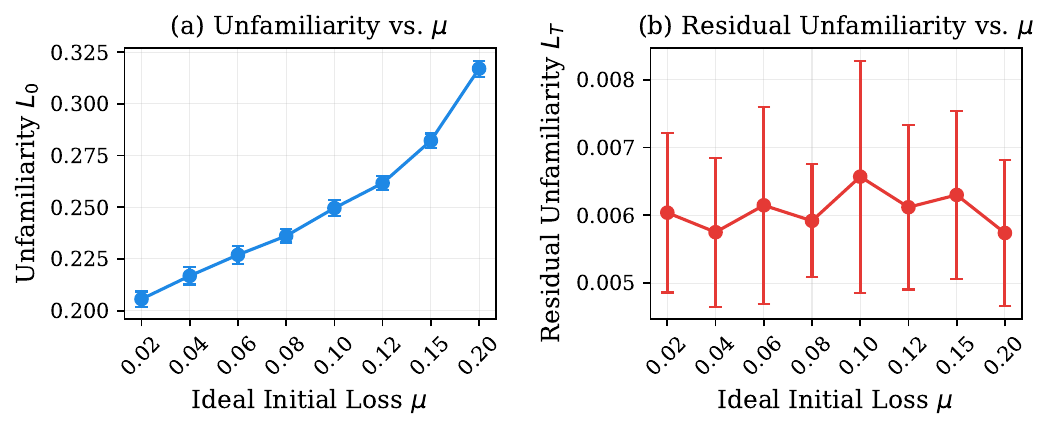}
\caption{Effect of ideal initial loss $\mu$ on generated image statistics ($\eta = 10$). \textbf{(a)}~Unfamiliarity $L_0$ grows linearly with $\mu$, so the complexity gate sets the level of unfamiliarity. \textbf{(b)}~Residual unfamiliarity $L_T$ stays near $0.006$ for every $\mu$.}
\label{fig:mu_analysis}
\end{figure}

\begin{figure}[h]
\centering
\includegraphics[width=\textwidth]{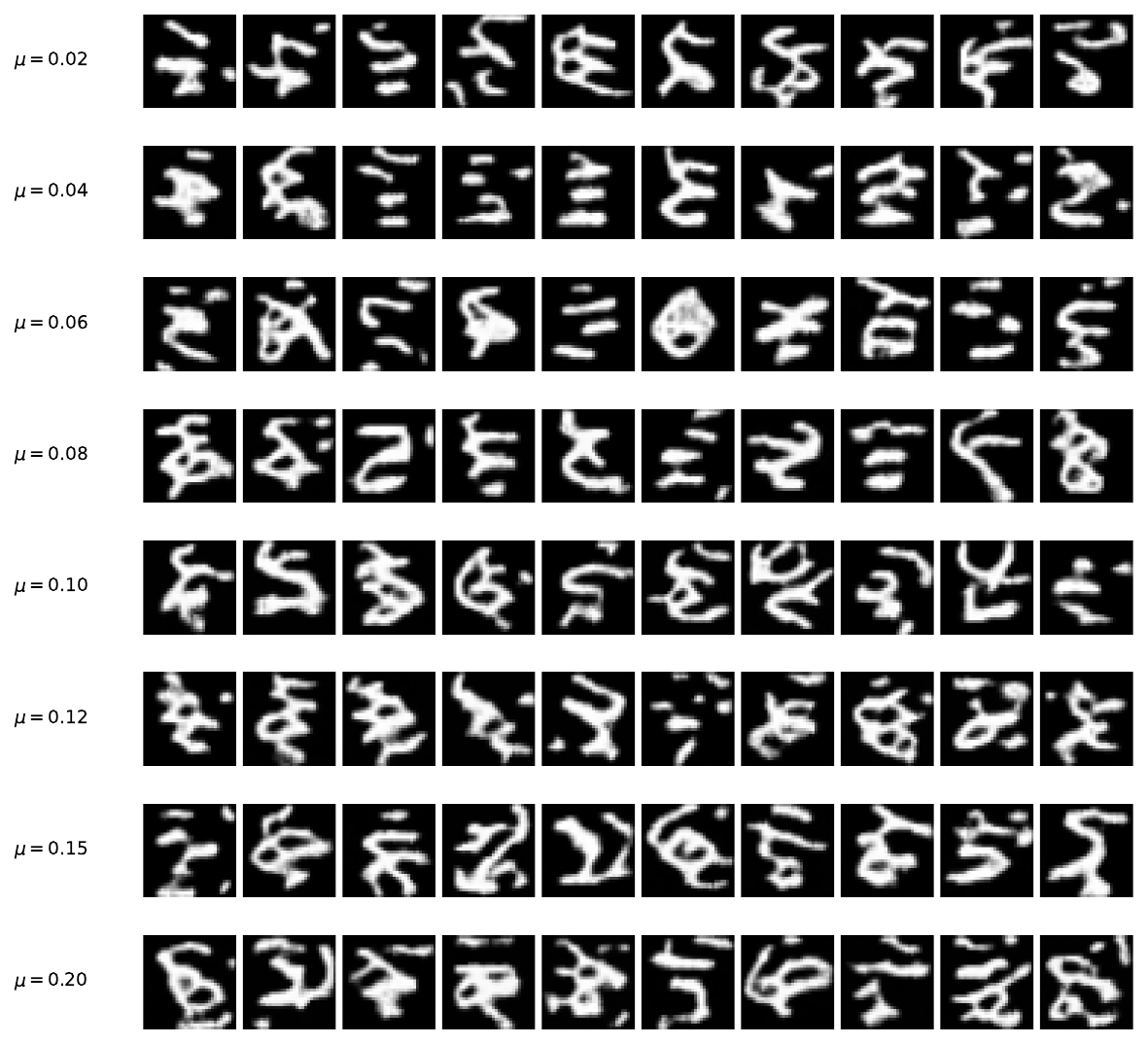}
\caption{Generated images across ideal initial loss $\mu$ values ($\eta = 10$), corresponding to the analysis in Figure~\ref{fig:mu_analysis}. Lower $\mu$ gives thinner, glyph-like forms, and higher $\mu$ gives thicker strokes that depart further from the training digits.}
\label{fig:mu_sweep}
\end{figure}

\begin{figure}[h]
\centering
\includegraphics[width=\textwidth]{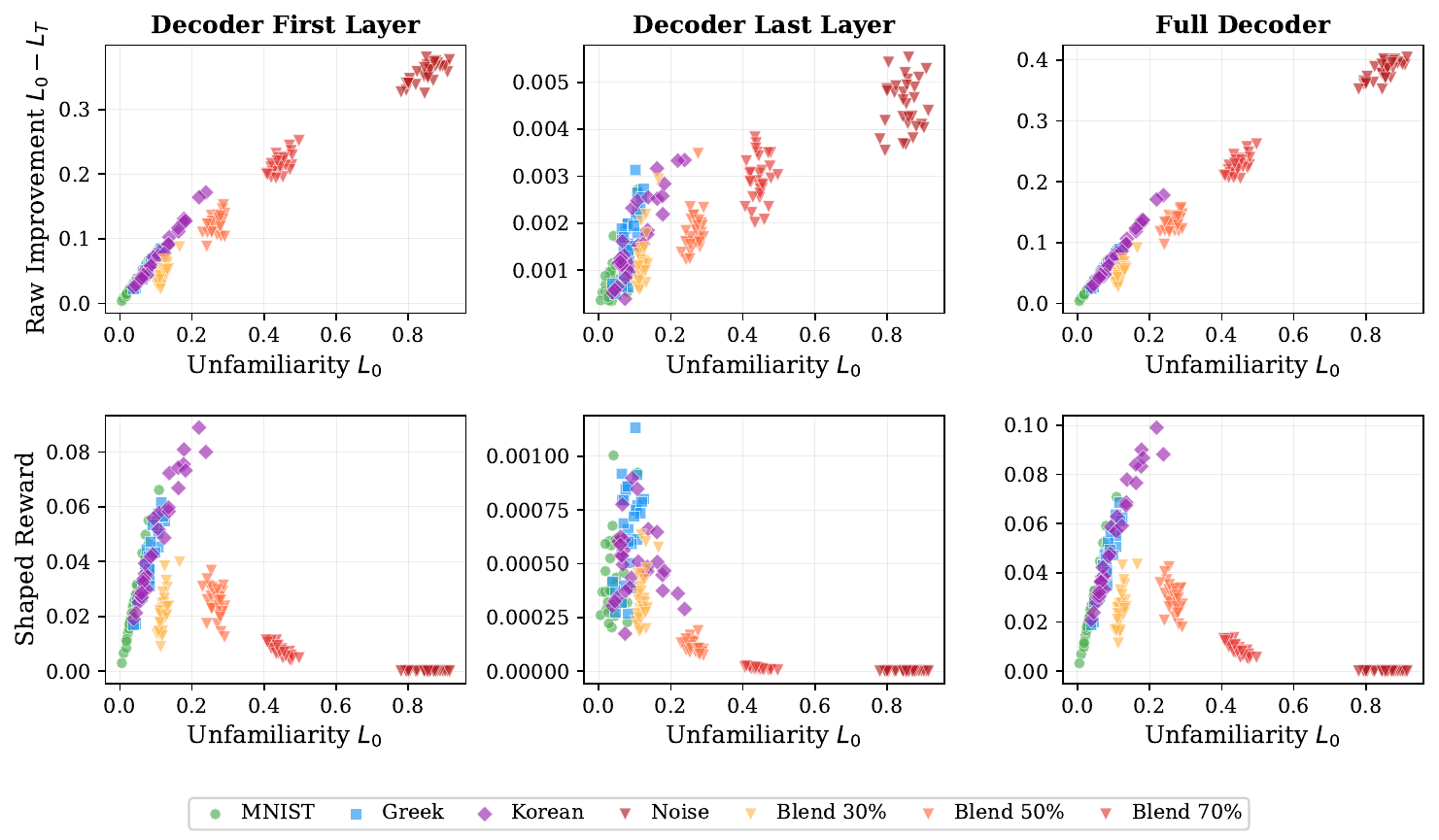}
\caption{Layer-specific learnability. The shaped reward when fine-tuning only individual decoder layers during the inner loop, against fine-tuning the full decoder (rightmost). Adapting only the first (high-level, semantic) layer recovers most of the full-decoder reward across categories, while adapting only the last (low-level, textural) layer yields near-zero improvement on every category.}
\label{fig:layer_ablation}
\end{figure}

This section gives MNIST samples for each baseline (Figure~\ref{fig:baseline_grid}), the effect of the guidance scale $\eta$ and the gate target $\mu$ (Figures~\ref{fig:scale_analysis}--\ref{fig:mu_sweep}), and the layer ablation (Figure~\ref{fig:layer_ablation}).

\section{Automatic Top-K Pipeline and Outputs Across More Prompts}
\label{sec:appendix_bon}

\begin{figure}[h!]
\centering
\includegraphics[width=0.7\textwidth]{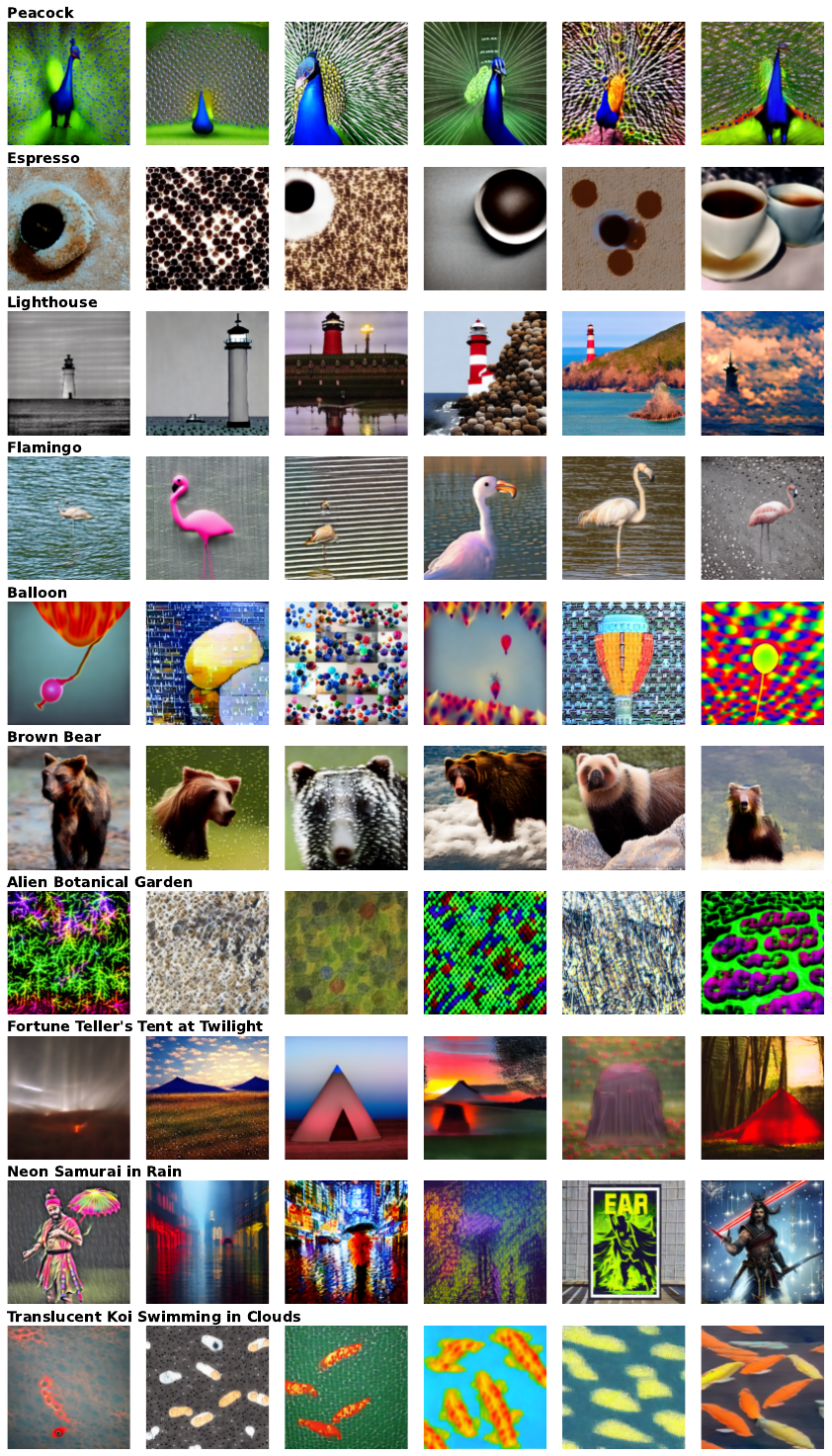}
\caption{Automatic top-K outputs from the Creator-Appraiser framework on 10 additional prompts not shown in the main text. Each row is one prompt; six columns are the random sample of top-8 by $\Delta\text{sim}$ within the relevance gate (seed fixed). No manual curation.}
\label{fig:appendix_bon_picks}
\end{figure}

The samples in Figure~\ref{fig:clip_lora_picks} were picked by hand. Figure~\ref{fig:appendix_bon_picks} shows samples picked by an automatic pipeline on 10 further prompts.

\paragraph{Generation.} For each new prompt, we run the same Creator-Appraiser pipeline as in Section~\ref{sec:natural_images}: a Stable Diffusion v1.5 Creator with the prompt embedding as the learnable variable, a frozen CLIP ViT-L/14 image encoder, and a LoRA inner loop on CLIP's text projection. We sweep LoRA rank $r \in \{4, 8\}$ and produce 100 samples per (prompt, rank) cell at outer-loop guidance $\eta = 10$, $T = 10$ inner-loop steps with learning rate $10^{-3}$, and 50 denoising steps.

\paragraph{Per-image scoring.} Every generated image is scored along three axes:
\begin{itemize}
\item $\text{sim}_0$, the frozen-CLIP image-text cosine similarity with the prompt (no LoRA). This is the inner-loop initial similarity.
\item $\Delta\text{sim} = \text{sim}_T - \text{sim}_0$, the similarity improvement after a fresh $T$-step LoRA inner loop on the saved image. This re-scores the image against the same predictor used during generation.
\item A Qwen2.5-VL-7B-Instruct relevance score~\citep{bai2025qwen2vl} on a 0--10 scale, prompted to judge \emph{whether the prompt's concept is recognizable in the image}, explicitly instructed not to lower the score for stylistic departure. This signal is independent of CLIP; the two encoders share no parameters.
\end{itemize}

\paragraph{Gating and ranking.} We pass an image through the gate if $\text{sim}_0 \ge 0.12$ AND the VLM relevance is $\ge 5$. The $\text{sim}_0$ gate rejects broken outputs, and the VLM gate checks relevance. Survivors are ranked by $\Delta\text{sim}$ in descending order; the top-$8$ are kept as candidate picks.

\paragraph{Sampling for the figure.} For each prompt we show a random 6 of the top 8, with a fixed seed, so the figure is not filled with near-copies of the single highest-$\Delta\text{sim}$ image.

\section{Reachability: DINOv2 Distance from the Vanilla Samples}
\label{sec:appendix_reachability}

We draw $N=1000$ vanilla samples per prompt, embed both vanilla samples and our samples with a frozen DINOv2 ViT-L/14~\citep{oquab2024dinov2}, $L_2$-normalize, and report nearest-neighbor cosine distances, mirroring the per-sample rarity construction of \citet{han2022rarity}. For one of our samples $x$, $d_V(x) = \min_{x' \in V} \big[1 - f(x)^\top f(x')\big]$ measures how far $x$ is from the vanilla set $V$; for a vanilla sample $x'$, $d_V(x') = \min_{x'' \in V \setminus \{x'\}} \big[1 - f(x')^\top f(x'')\big]$ is the typical distance within $V$.

Table~\ref{tab:dv_per_prompt} gives the per-prompt distances for our method and the diversity baselines. At $r{=}8$, the ratio of our distance to the intra-vanilla distance ranges from $1.9\times$ (neon samurai in rain) to $17\times$ (peacock).

This reachability metric is not itself a creativity score. DINOv2 distance measures atypicality relative to vanilla sampling, while the Appraiser's inner-loop reward supplies the learnability criterion and the frozen-CLIP $\text{sim}_0$ and Qwen2.5-VL relevance check tie the sample to the prompt. We read the three together as ``unfamiliar but memorable''.

\paragraph{Selection controls.} The relevance gate keeps fewer samples at higher rank, so the gated medians are computed over smaller subsets. Table~\ref{tab:selection_controls} repeats the analysis without the gate and with a fixed count, keeping the 13 samples per prompt with the highest Qwen2.5-VL relevance score at every rank. The distance increases with rank under all three protocols, and the ungated distances are higher than the gated ones, so the gate makes the reported numbers conservative. The atypical tail of the vanilla pool stays on-prompt, and only the single most atypical of 1000 vanilla samples approaches the distance of our gated samples at $r{=}8$.

\begin{table}[h]
\centering
\caption{DINOv2 distance to the vanilla pool, median over 20 prompts (18 for gated $r{=}32$). Top: selection protocols for our samples. Bottom: vanilla samples by atypicality.}
\label{tab:selection_controls}
\small
\begin{tabular}{@{}lcccc@{}}
\toprule
 & $r{=}4$ & $r{=}8$ & $r{=}16$ & $r{=}32$ \\
\midrule
Gated (relevance $\geq 5$) & 0.41 & 0.53 & 0.55 & 0.66 \\
Ungated (all 100 samples) & 0.57 & 0.73 & 0.81 & 0.84 \\
Top 13 by relevance & 0.40 & 0.53 & 0.65 & 0.74 \\
\midrule
Vanilla & Typical & Top 10\% & Top 5\% & Most atypical \\
\midrule
Distance & 0.11 & 0.17 & 0.21 & 0.52 \\
Relevance rate & 0.96 & 0.96 & 0.95 & 0.85 \\
\bottomrule
\end{tabular}
\end{table}

\paragraph{Compute cost.} On one H200 GPU with 50 denoising steps, our method at $r{=}8$ with $T{=}10$ inner-loop steps takes 7.7\,s per image, against 0.86\,s for vanilla Stable Diffusion, about $9\times$ more. We time each method at 2 and 20 images and divide the difference by 18, which removes model loading. The cost is expected to grow linearly with the number of inner-loop and denoising steps.
\begin{table}[t]
\centering
\caption{Baselines on 20 prompts. Relevance rate: fraction with Qwen2.5-VL relevance $\geq 5$. NN distance and $\text{sim}_0$: medians over prompts on those samples, with distance to the 1000 vanilla samples (vanilla: to the other 999). ProCreate uses CLIP features in place of the original DreamSim features. Its $\gamma{=}1000$ row and our $r{=}32$ row cover 19 and 18 prompts.}
\label{tab:baselines_full}
\small
\begin{tabular}{@{}lccc@{}}
\toprule
Method & Relevance $\geq 5$ rate & NN distance & $\text{sim}_0$ \\
\midrule
Vanilla (CFG 7.5) & 0.96 & 0.11 & 0.27 \\
CFG scale $s \in \{3, 5, 10\}$ & 0.94--0.97 & 0.06--0.17 & 0.26--0.27 \\
CFG scale $s = 1$ & 0.64 & 0.47 & 0.23 \\
CFG scale $s = 0.5$ & 0.17 & 0.66 & 0.20 \\
Negative prompt (4 variants) & 0.95--0.96 & 0.10--0.12 & 0.26 \\
ProCreate~\citep{lu2024procreate}, $\gamma = 250$ & 0.73 & 0.38 & 0.21 \\
ProCreate, $\gamma = 500$ & 0.28 & 0.58 & 0.18 \\
ProCreate, $\gamma = 1000$ & 0.08 & 0.72 & 0.15 \\
\midrule
Ours, $r = 4$ & 0.62 & 0.41 & 0.23 \\
Ours, $r = 8$ & 0.27 & 0.53 & 0.20 \\
Ours, $r = 16$ & 0.14 & 0.55 & 0.18 \\
Ours, $r = 32$ & 0.07 & 0.66 & 0.16 \\
\bottomrule
\end{tabular}
\end{table}

\begin{figure}[h]
\centering
\includegraphics[width=0.8\textwidth]{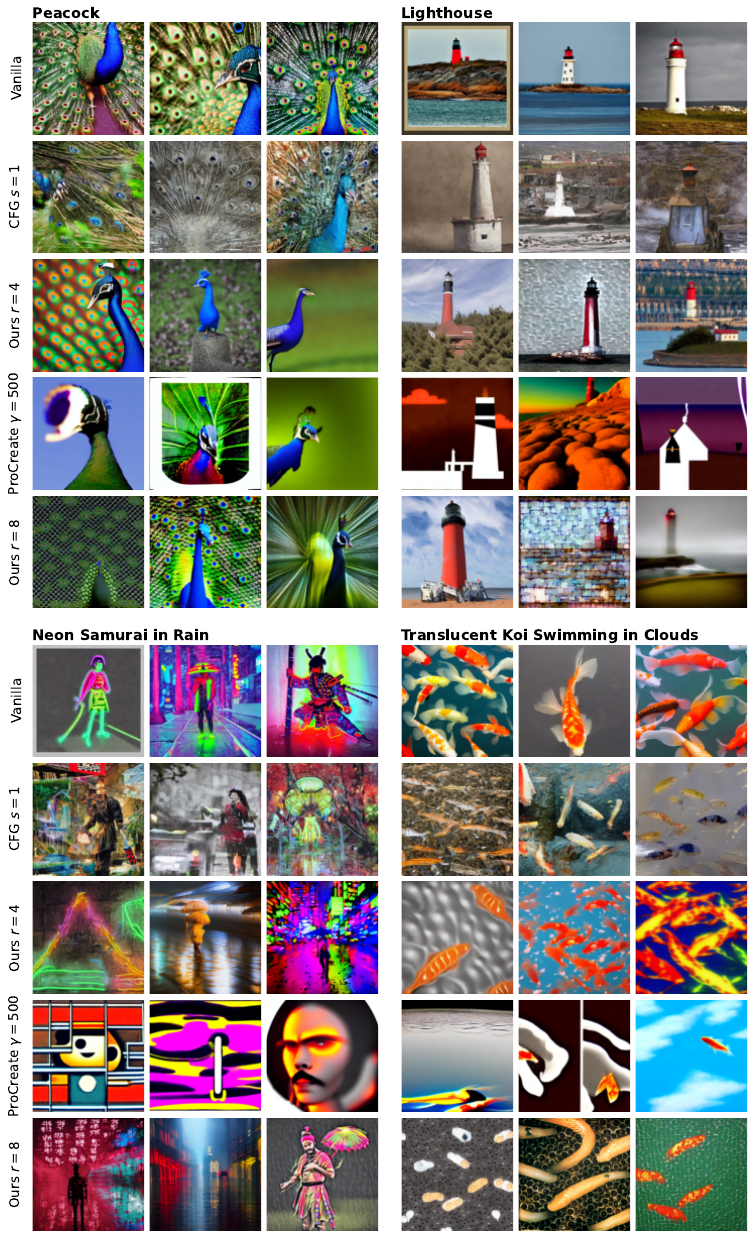}
\caption{Samples of the diversity baselines and ours on four prompts. CFG $s{=}1$ is paired with ours at $r{=}4$ and ProCreate $\gamma{=}500$ with ours at $r{=}8$, at matched relevance rate (Table~\ref{tab:baselines_full}). Each cell is a random draw of samples with Qwen2.5-VL relevance $\geq 5$. Vanilla (CFG 7.5) samples are random and ungated.}
\label{fig:appendix_baseline_compare}
\end{figure}

\paragraph{Baseline samples.} Figure~\ref{fig:appendix_baseline_compare} shows samples of the diversity baselines next to ours, paired at matched relevance rate. CFG $s{=}1$ samples are desaturated and blurry, ProCreate $\gamma{=}500$ samples often become flat graphic shapes, and our samples vary in color, texture, and composition.

\begin{figure}[h]
\centering
\includegraphics[width=0.9\textwidth]{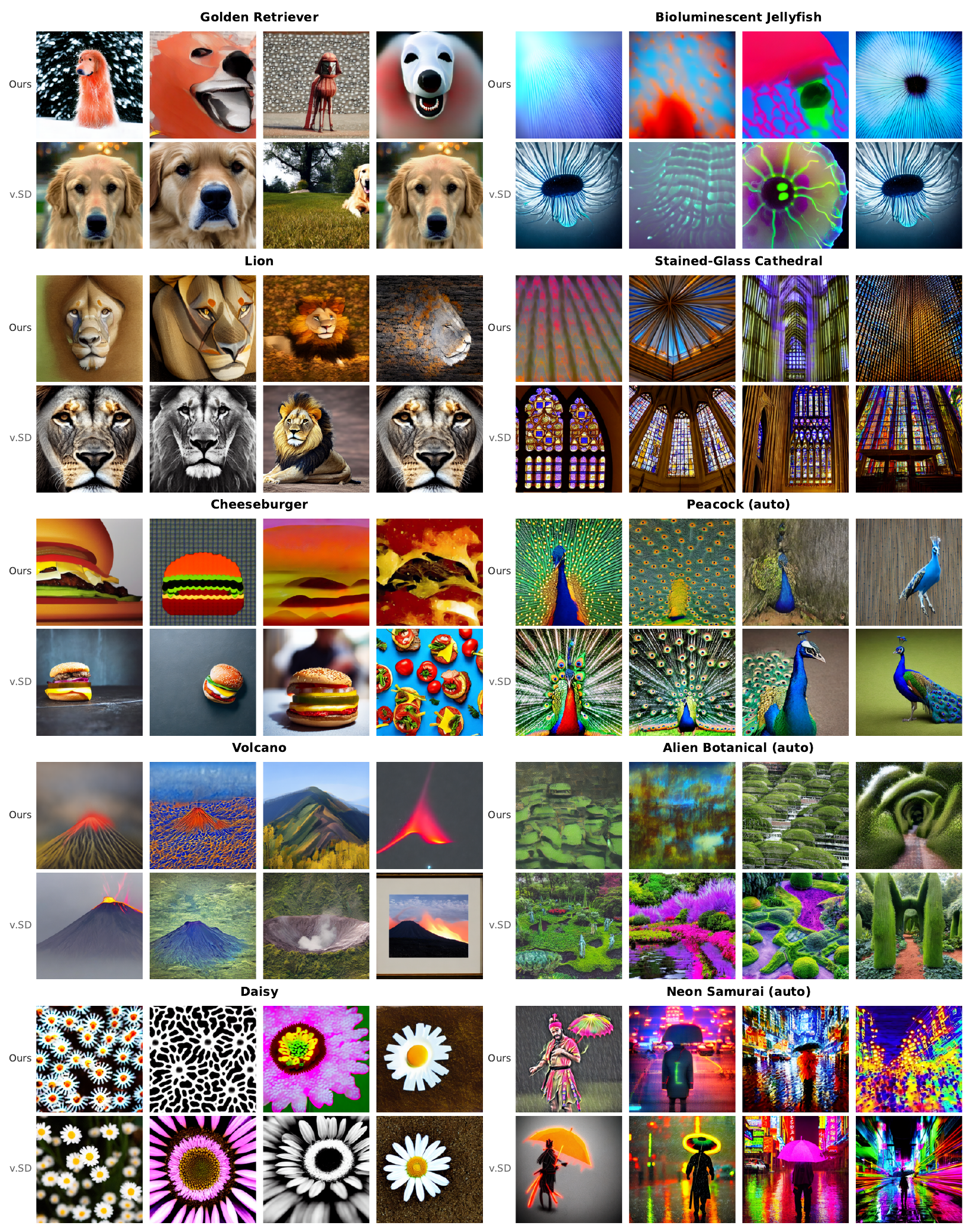}
\caption{Each picked Creator-Appraiser sample (top of each pair) shown alongside its DINOv2-nearest neighbor (bottom of each pair) selected from 1000 vanilla Stable Diffusion samples. Top 7 prompts are the manually curated picks from Figure~\ref{fig:clip_lora_picks}; bottom 3 are automatic picks for three of the prompts in Figure~\ref{fig:appendix_bon_picks} (peacock, an alien botanical garden, neon samurai in rain).}
\label{fig:appendix_paired_comparison}
\end{figure}

Figure~\ref{fig:appendix_paired_comparison} pairs each picked sample with its DINOv2-nearest vanilla counterpart from the 1000-sample set. In the rightmost pair of the Daisy row of Figure~\ref{fig:appendix_paired_comparison}, our sample matches the vanilla counterpart's lighting, background tone, and even rough framing, but the flower head itself adopts a different geometry, with petals that read as ink-stroke abstractions rather than the photographic daisy that vanilla converges to.

\begin{table}[h]
\centering
\footnotesize
\setlength{\tabcolsep}{3.5pt}
\caption{Per-prompt mean DINOv2 nearest-neighbor distance to 1000 vanilla samples, on samples with Qwen2.5-VL relevance $\geq 5$. Intra-V: distance of each vanilla sample to the other 999. PC: ProCreate. --: no sample passes.}
\label{tab:dv_per_prompt}
\begin{tabular}{@{}lc|ccccc|cccc@{}}
\toprule
& & \multicolumn{5}{c|}{Baselines} & \multicolumn{4}{c}{Ours, LoRA rank} \\
Prompt & Intra-V & CFG $1$ & CFG $0.5$ & PC $250$ & PC $500$ & PC $1000$ & $4$ & $8$ & $16$ & $32$ \\
\midrule
\multicolumn{11}{@{}l}{\textit{ImageNet class prompts}} \\
balloon & 0.20 & 0.53 & 0.51 & 0.37 & 0.49 & 0.61 & 0.38 & 0.51 & 0.49 & 0.57 \\
brown bear & 0.04 & 0.43 & 0.69 & 0.15 & 0.42 & 0.87 & 0.22 & 0.43 & 0.54 & -- \\
castle & 0.28 & 0.65 & 0.75 & 0.57 & 0.81 & 0.85 & 0.57 & 0.66 & 0.69 & 0.80 \\
cheeseburger & 0.08 & 0.33 & 0.53 & 0.35 & 0.51 & 0.84 & 0.21 & 0.38 & 0.36 & 0.42 \\
daisy & 0.16 & 0.55 & 0.72 & 0.43 & 0.70 & 0.79 & 0.47 & 0.61 & 0.74 & 0.78 \\
espresso & 0.10 & 0.42 & 0.58 & 0.39 & 0.58 & 0.63 & 0.41 & 0.58 & 0.74 & 0.68 \\
flamingo & 0.04 & 0.37 & 0.68 & 0.18 & 0.46 & 0.72 & 0.10 & 0.19 & 0.22 & 0.20 \\
golden retriever & 0.10 & 0.50 & 0.74 & 0.25 & 0.42 & 0.59 & 0.42 & 0.50 & 0.53 & 0.84 \\
jellyfish & 0.07 & 0.49 & 0.68 & 0.32 & 0.62 & 0.78 & 0.36 & 0.59 & 0.52 & 0.69 \\
lighthouse & 0.14 & 0.48 & 0.80 & 0.34 & 0.58 & 0.68 & 0.32 & 0.33 & 0.40 & 0.47 \\
lion & 0.03 & 0.22 & 0.58 & 0.14 & 0.38 & -- & 0.17 & 0.38 & 0.38 & 0.40 \\
peacock & 0.01 & 0.15 & 0.66 & 0.12 & 0.31 & 0.72 & 0.12 & 0.25 & 0.42 & 0.51 \\
volcano & 0.07 & 0.30 & 0.52 & 0.29 & 0.57 & 0.75 & 0.25 & 0.27 & 0.42 & 0.70 \\
\midrule
\multicolumn{11}{@{}l}{\textit{Open-ended prompts}} \\
alien botanical garden & 0.22 & 0.47 & 0.60 & 0.57 & 0.70 & 0.79 & 0.66 & 0.78 & 0.79 & 0.80 \\
ancient library lit by fireflies & 0.20 & 0.55 & 0.66 & 0.66 & 0.75 & 0.76 & 0.62 & 0.70 & 0.77 & 0.80 \\
bioluminescent jellyfish & 0.12 & 0.47 & 0.59 & 0.46 & 0.60 & 0.66 & 0.53 & 0.61 & 0.63 & 0.64 \\
fortune teller's tent at twilight & 0.19 & 0.50 & 0.66 & 0.39 & 0.59 & 0.69 & 0.46 & 0.54 & 0.65 & 0.60 \\
neon samurai in rain & 0.28 & 0.48 & 0.50 & 0.47 & 0.56 & 0.59 & 0.48 & 0.53 & 0.56 & 0.65 \\
stained-glass cathedral interior & 0.12 & 0.45 & 0.67 & 0.52 & 0.61 & 0.66 & 0.45 & 0.58 & 0.75 & -- \\
translucent koi swimming in clouds & 0.05 & 0.42 & 0.68 & 0.39 & 0.62 & 0.78 & 0.50 & 0.60 & 0.72 & 0.77 \\
\midrule
\textbf{Median} & 0.11 & 0.47 & 0.66 & 0.38 & 0.58 & 0.72 & 0.41 & 0.53 & 0.55 & 0.66 \\
\bottomrule
\end{tabular}
\end{table}

\begin{figure}[h]
\centering
\includegraphics[width=0.55\textwidth]{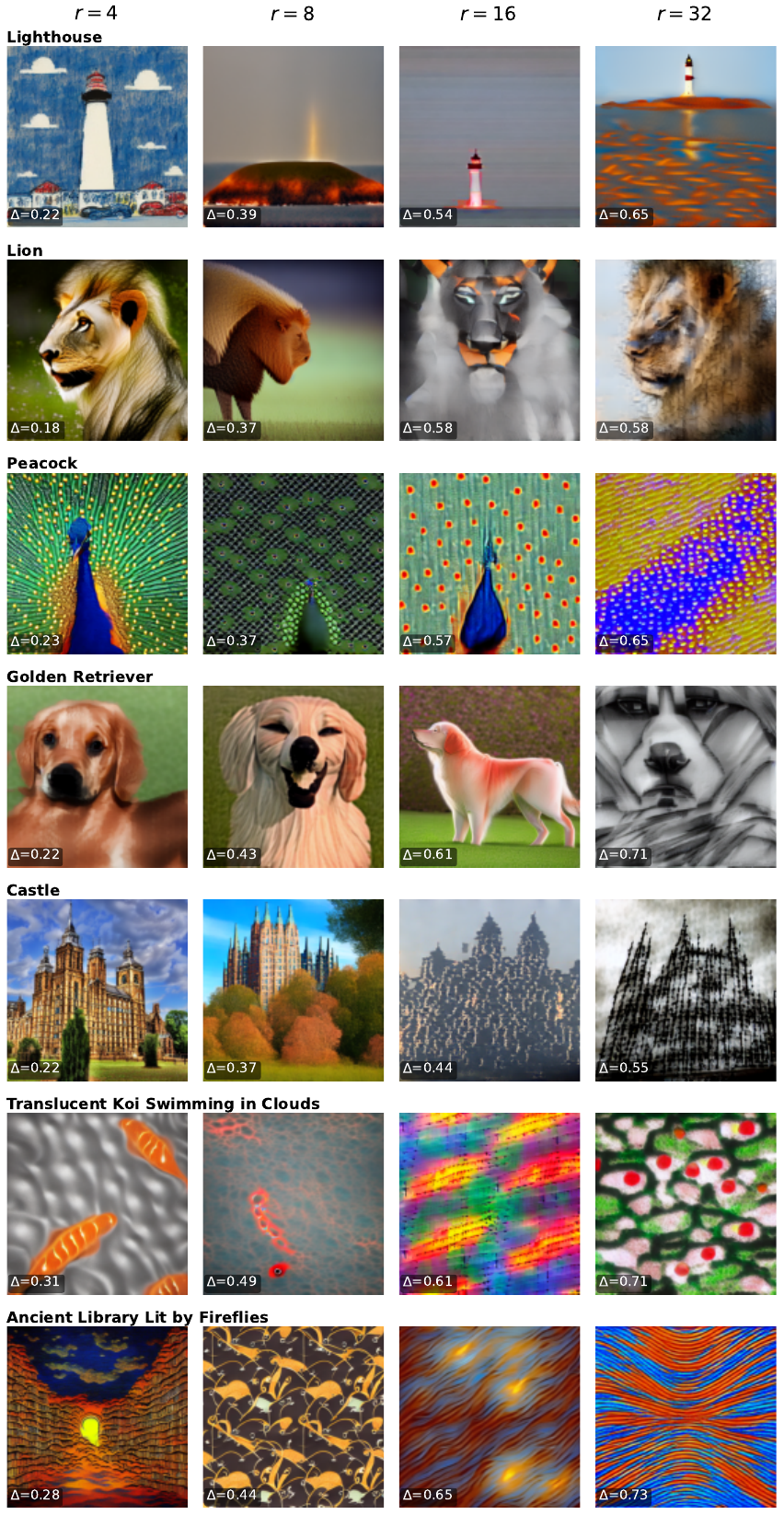}
\caption{Effect of LoRA rank: Per prompt, one hand-picked sample at LoRA rank $r \in \{4, 8, 16, 32\}$. As rank grows the inner LoRA gains capacity to assimilate increasingly out-of-distribution outputs, so the meta-gradient is free to push the diffusion sample further from the prompt.}
\label{fig:appendix_rank_gradient}
\end{figure}

\paragraph{Effect of LoRA rank.} Figure~\ref{fig:appendix_rank_gradient} shows each prompt rendered at $r \in \{4, 8, 16, 32\}$, with one sample per cell, picked by hand from samples ranked by $\Delta\text{sim}$. Low-rank cells stay close to the photograph, and high-rank cells move toward stylization, abstraction, or texture. In the median over prompts, the distance to vanilla and the inner-loop reward $\Delta\text{sim}$ both grow with rank (Tables~\ref{tab:dv_per_prompt} and~\ref{tab:natural_image_quant_full}).

\section{Per-Prompt Natural Image Results}

Table~\ref{tab:natural_image_quant_full} gives the per-prompt values behind Figure~\ref{fig:rank_tradeoff}, $\text{sim}_0$ for vanilla Stable Diffusion, and $\text{sim}_0$ and $\Delta\text{sim}$ for our method at each rank.

\begin{table}[h]
\centering
\small
\caption{Per-prompt natural image results, restricted to our samples passing the Qwen2.5-VL relevance gate ($\geq 5$, same filter as Table~\ref{tab:dv_per_prompt}). Cells in the rank columns are $\text{sim}_0\,/\,\Delta\text{sim}$, where $\text{sim}_0$ is frozen-CLIP image-text cosine before adaptation (concept fidelity) and $\Delta\text{sim}$ is the inner-loop reward. --: no sample passes.}
\begin{tabular}{lccccc}
\toprule
& Vanilla SD & \multicolumn{4}{c}{Ours\quad ($\text{sim}_0$ / $\Delta\text{sim}$)} \\
\cmidrule(lr){3-6}
Prompt & $\text{sim}_0$ & $r{=}4$ & $r{=}8$ & $r{=}16$ & $r{=}32$ \\
\midrule
\multicolumn{6}{l}{\textit{ImageNet class prompts}} \\
balloon                                & 0.26 & 0.24 / 0.20 & 0.21 / 0.36 & 0.21 / 0.54 & 0.19 / 0.61 \\
brown bear                             & 0.27 & 0.25 / 0.20 & 0.20 / 0.37 & 0.18 / 0.55 & -- \\
castle                                 & 0.23 & 0.22 / 0.19 & 0.19 / 0.35 & 0.18 / 0.52 & 0.17 / 0.66 \\
cheeseburger                           & 0.26 & 0.23 / 0.22 & 0.19 / 0.39 & 0.21 / 0.57 & 0.15 / 0.70 \\
daisy                                  & 0.27 & 0.23 / 0.20 & 0.21 / 0.36 & 0.19 / 0.52 & 0.19 / 0.59 \\
espresso                               & 0.24 & 0.22 / 0.20 & 0.19 / 0.37 & 0.21 / 0.49 & 0.15 / 0.67 \\
flamingo                               & 0.29 & 0.27 / 0.20 & 0.25 / 0.35 & 0.24 / 0.51 & 0.20 / 0.62 \\
golden retriever                       & 0.28 & 0.22 / 0.23 & 0.22 / 0.40 & 0.18 / 0.60 & 0.17 / 0.68 \\
jellyfish                              & 0.27 & 0.23 / 0.22 & 0.19 / 0.39 & 0.18 / 0.55 & 0.15 / 0.65 \\
lighthouse                             & 0.25 & 0.24 / 0.20 & 0.23 / 0.36 & 0.22 / 0.52 & 0.21 / 0.62 \\
lion                                   & 0.26 & 0.23 / 0.19 & 0.21 / 0.35 & 0.20 / 0.53 & 0.21 / 0.61 \\
peacock                                & 0.29 & 0.27 / 0.23 & 0.24 / 0.39 & 0.19 / 0.56 & 0.16 / 0.66 \\
volcano                                & 0.24 & 0.25 / 0.21 & 0.24 / 0.36 & 0.21 / 0.54 & 0.22 / 0.64 \\
\midrule
\multicolumn{6}{l}{\textit{Open-ended prompts}} \\
alien botanical garden                 & 0.27 & 0.20 / 0.20 & 0.17 / 0.36 & 0.15 / 0.54 & 0.14 / 0.62 \\
ancient library lit by fireflies       & 0.28 & 0.20 / 0.26 & 0.15 / 0.44 & 0.12 / 0.60 & 0.11 / 0.70 \\
bioluminescent jellyfish               & 0.28 & 0.17 / 0.25 & 0.16 / 0.43 & 0.10 / 0.66 & 0.05 / 0.82 \\
fortune teller's tent at twilight      & 0.25 & 0.22 / 0.25 & 0.21 / 0.41 & 0.17 / 0.59 & 0.15 / 0.68 \\
neon samurai in rain                   & 0.34 & 0.21 / 0.26 & 0.17 / 0.44 & 0.13 / 0.60 & 0.13 / 0.66 \\
stained-glass cathedral interior       & 0.25 & 0.23 / 0.24 & 0.19 / 0.40 & 0.16 / 0.55 & -- \\
translucent koi swimming in clouds     & 0.27 & 0.21 / 0.26 & 0.18 / 0.46 & 0.13 / 0.65 & 0.12 / 0.71 \\
\midrule
\textbf{Mean}                          & \textbf{0.27} & \textbf{0.23 / 0.22} & \textbf{0.20 / 0.39} & \textbf{0.18 / 0.56} & \textbf{0.16 / 0.66} \\
\textbf{Median}                        & \textbf{0.27} & \textbf{0.23 / 0.21} & \textbf{0.20 / 0.38} & \textbf{0.18 / 0.55} & \textbf{0.16 / 0.66} \\
\bottomrule
\end{tabular}
\label{tab:natural_image_quant_full}
\end{table}

\end{document}